\documentclass[10pt,twocolumn,twoside]{IEEEtran}
\usepackage{amsfonts,amsmath, amssymb,euscript,amsbsy, mathtools} % mathtools for \DeclarePairedDelimiterX
\allowdisplaybreaks % from ansmath
\usepackage{graphicx, color, adjustbox, float}
\usepackage{authblk} % \affil (command for authors affiliation)
\usepackage{pgffor} % \foreach (looping through some elements)
\usepackage{bbm} % \mathbbm for vector of ones
\usepackage{cases} % numcases and subnumcases environments for math cases
\usepackage{bm} % \bm bold symbols
\usepackage{array}	% control the vertical alignment of tables
\usepackage{pifont} % for check and xmark
\newcommand{\cmark}{\ding{51}}%
\newcommand{\xmark}{\ding{55}}%
\usepackage[noadjust]{cite}
\usepackage{epstopdf,booktabs,multirow}
\usepackage[hidelinks]{hyperref}	% clickable references
\graphicspath{{./Figs/}}
\usepackage[lined,boxed,linesnumbered, ruled,vlined]{algorithm2e}
\newcommand{\minitab}[2][l]{\begin{tabular}{#1}#2\end{tabular}}
\begin{document}
%--------------------------------------------------------------------------------------------------
%								Title, authors, and affiliations
%--------------------------------------------------------------------------------------------------
\title{Privacy-Preserving Collaborative Split Learning Framework for Smart Grid Load Forecasting}

\author{{Asif~Iqbal, Prosanta~Gope,~\textit{Senior Member, IEEE}, and Biplab~Sikdar,~\textit{Senior Member, IEEE}}%
\thanks{Asif~Iqbal and Biplab~Sikdar are with the Department of Electrical and Computer Engineering, National University of Singapore, Singapore. E-mails: aiqbal, bsikdar@nus.edu.sg.}%
\thanks{Prosanta~Gope is with the Department of Computer Science at the University of Sheffield, United Kingdom. E-mail: p.gope@sheffield.ac.uk}%}%
\thanks{This work was supported in part by the Asian Institute of Digital Finance (AIDF) under Grant A-0003504-09-00.}
\thanks{\textbf{Corresponding author:} Dr. Prosanta~Gope.}}
%--------------------------------------------------------------------------------------------------
%								Paper header
%--------------------------------------------------------------------------------------------------
\markboth{IEEE Transactions on Dependable and Secure Computing}%
{Iqbal \MakeLowercase{\textit{et al.}}: Collaborative Split Learning Framework for Smart Grid Load Forecasting}

%--------------------------------------------------------------------------------------------------
%									Make title, abstract, keywords
\maketitle

\begin{abstract}
Accurate load forecasting is crucial for energy management, infrastructure planning, and demand-supply balancing. The availability of smart meter data has led to the demand for sensor-based load forecasting. 
Conventional ML allows training a single global model using data from multiple smart meters requiring data transfer to a central server, raising concerns for network requirements, privacy, and security.
{To alleviate this issue, we} propose a split learning-based framework for load forecasting. We split a deep neural network model into two parts, one for each Grid Station (GS) responsible for an entire neighbourhood's smart meters and the other for the Service Provider (SP). Instead of sharing their data, client smart meters use their respective GSs' model split for forward passes and only share their activations with the GS.
Under this framework, each GS is responsible for training a personalized model split for their respective neighbourhoods, whereas
the SP can train a single global or personalized model for each GS. 
Experiments show that the proposed models match or exceed a centrally trained model's performance and generalize well.
Privacy is analyzed by assessing information leakage between data and shared activations of the GS model split.

%Additionally, differential privacy enhances local data privacy while examining its impact on performance. A transformer model is used as our base learner. 
\end{abstract}


\IEEEpeerreviewmaketitle
% Manuscript files included below
%======================================================================================================
\section{Introduction} \label{sec:Intro}
Electricity load forecasting is crucial for energy management systems as it enables planning for power infrastructure upgrades, demand and supply balancing, and power generation scheduling in response to renewable energy fluctuations. Accurate load forecasting can also result in significant cost savings. In 2016, Xcel Energy saved \$2.5 million by reducing their load forecasting error from 15.7\% to 12.2\% \cite{Notton2018Foi}.

Sensor-based approaches for electricity load forecasting use historical load traces from smart meters and meteorological data to train machine learning (ML) models. ML models for load forecasting can be trained through {localized} or {centralized} methods. Localized training entails developing a dedicated model for each smart meter, facilitating client-level load forecasting. Conversely, centralized training involves building a single model using aggregated data from multiple clients to forecast the load for an entire area \cite{Arif2020Ela}. However, transferring data directly from clients' premises to a centralized server imposes a heavy communication load and raises significant privacy and security concerns \cite{Truong2020PPi}.
% Although both methods perform well, their centralized framework leads to a high communications load and serious privacy and security concerns \cite{Truong2020PPi}. 
For instance, high-resolution smart meter data might disclose when someone is at home, their daily routines, and even specific activities. Moreover, their load signatures can identify certain electrical devices or appliances. For example, the use of medical equipment, home security systems, or specialized machinery can be inferred from the load data \cite{Reinhardt2012electric}. Safeguarding this information is of utmost importance, as it protects an individual's privacy and ensures compliance with stringent data regulations, such as the European Union General Data Protection Regulation (GDPR) \cite{Hoofnagle2019GDPA}. 
In addition, the growing adoption of smart meters renders the practice of training individual models for each customer, whether locally or centrally, increasingly impractical from both computational and financial standpoints.

In order to alleviate these problems, decentralized deep learning methods like \textit{federated learning} (FL) \cite{Bonawi2019Tfl} and \textit{split learning} (SL/SplitNN) \cite{Vepako2018Slf} have been proposed. These methods decouple the requirement of training an ML model on locally/centrally available data by enabling a group of data holders to train an ML model collaboratively without sharing their private data. In FL, a server contains a global ML model shared among multiple clients. During training, each client receives a copy of the global ML model, generates a model update by improving its private data and sends the updated model back to the server. 
% Once the server has received enough model updates from the clients, it 
The server then performs aggregation and updates the global model in some way, usually via weighted averaging, and sends the updated model back to each client. 
In SplitNN, the ML model is split into two parts, one remains on the client's side and the other on the server's side. The client performs the forward pass on its side and shares the outputs with the server, which continues the forward propagation and computes the loss. The gradients are sent back to the client to complete an update step. Clients can choose any ML architecture, as the server has no control over it.
In both FL and SplitNN, clients do not share their private data with anyone.  
These decentralized learning approaches have resulted in a major paradigm shift from an expensive central ML system to utilizing various distributed computational resources.  
%%%%%%%%%%%%%%%%%%%%%%%%%%%%%%%%%%%%%%%%%%%%%%%%
\subsection{Related works} \label{sec:RelWorks}
This section reviews the recent ML methods proposed for load forecasting, followed by studies that use FL and SL for distributive load forecasting.
%%%%%%%%%%%%%%%%%%%%%%%%%%%%%%%%%%%%%%%%%%%%%%%%%%
\subsubsection{Load Forecasting} \label{sec:LoadFore}
Although the load forecasting problem is not new and several methods have been developed for it \cite{Sehova2020Dlf}, we focus on recently proposed deep learning (DL) techniques which have been dominant in sensor-based forecasting \cite{Arif2020Ela}. Among these DL architectures, recurrent neural networks (RNN) are well suited for learning the temporal patterns present in smart meter data and have been shown to outperform classical statistical and other ML approaches \cite{Arif2020Ela}. 

Authors in \cite{Sehova2020Dlf} have used the attention mechanism to develop a Sequence to Sequence RNN (S2S RNN) for load forecasting using two RNNs. 
% Using a dual network approach, they use two RNNs as encoder and decoder to learn input to output sequence mapping. 
Their use of an attention mechanism aids in capturing the long-term dependencies present in the load traces by improving the link between both RNNs.
In \cite{Tian2019Sbc}, authors used S2S RNN to perform load forecasting for several clients via Similarity Based Chained Transfer Learning (SBCTL), where they train a model for a single client traditionally while the other clients utilize transfer learning to build upon the already trained model. In \cite{Fekri2021Dlf}, authors present an online adaptive RNN model to train the model as new data arrives continuously. They use a Bayesian Normalized LSTM (BNLSTM) as their base learner and use an online Bayesian optimizer to update model weights online. Similarly, the authors in \cite{Ryu2023QM} propose a multi-layer perceptron mixer structure to perform 24-hour-ahead forecasting.
% Their proposed method was shown to outperform several other online learning methods as well as the conventional LSTM model.
% \begin{figure}
%     \centering
%     \includegraphics[width=0.95\columnwidth]{Transformer.pdf}
%     \caption{Recreation of the original Transformer architecture \cite{Vaswan2017Aia}.}
%     \label{fig:VanTransf}
% \end{figure}

% Despite the excellent performance of the above-mentioned RNN-based methods, it is well known that these methods suffer from the problem of vanishing (exploding) gradients where the gradient gets smaller and smaller with each layer until it is too small to affect the deepest layers. This results in performance deterioration once the sequence under consideration is long. 
Following an excellent performance in computer vision (CV) \cite{Rao2021Gfn} and natural language processing (NLP) community \cite{Vaswan2017Aia}, the Transformer architecture \cite{Vaswan2017Aia} 
% (shown in Fig. \ref{fig:VanTransf}) 
has recently been employed to capture long-term dependencies in time-series forecasting problems \cite{Wu2021Autoformer, Zhou2021Informer, Zhou2022FED}. Instead of working with a single time point at a time (as in RNN), transformer models perform sequence-to-sequence (instead of one sample ahead) forecasting using an encoder-decoder architecture. At the core of transformers, there are self-attention and cross-attention mechanisms which, in vanilla transformer \cite{Vaswan2017Aia}, 
% (see Fig. \ref{fig:VanTransf}), 
use a point-wise connected matrix leading to a quadratic computational complexity $\mathcal{O}(N^2)$ w.r.t. the input sequence size.

For the time-series forecasting problem, the quadratic complexity of the vanilla transformer is prohibitive. Thus, various modifications to the attention mechanism have been proposed to reduce its complexity.
% , various methods have been proposed to modify the attention matrix with pre-defined patterns to low. 
Authors in \cite{Li2019Logtrans} employ log-sparse attention to bring the complexity down to $\mathcal{O}(N\log^2 N)$. 
In \cite{Zhou2021Informer}, Informer architecture is proposed, which uses KL-divergence based ProbSparse self-attention mechanism and a distilling operation to reduce the complexity to $\mathcal{O}(N\log N)$. Authors in \cite{Wu2021Autoformer} propose Autoformer, which replaces the canonical attention with an auto-correlation block to achieve sub-series level attention with $\mathcal{O}(N\log N)$ complexity. 
% They use Fast Fourier transform (FFT) and top-k mode selection in the auto-correlation matrix to achieve $\mathcal{O}(N\log N)$ complexity. 
% In \cite{Zhu2021Ht1}, authors generate a sparse approximation of the attention matrix using a hierarchical pattern, leading to $\mathcal{O}(N)$ complexity. 
In FEDformer \cite{Zhou2022FED}, similar to \cite{Wu2021Autoformer}, authors replace the canonical attention with an attention mechanism implemented in the frequency domain (using FFT or wavelet transform). They perform low-rank approximation in the frequency domain and use the mixture of experts' decomposition to separate short-term and long-term patterns, leading to linear complexity $\mathcal{O}(N)$. 
Drawing on the comprehensive performance evaluations reported in FEDformer \cite{Zhou2022FED} and Autoformer \cite{Wu2021Autoformer}, these methods outperform recently proposed transformer-based, LSTM-based, and statistical-based approaches in both univariate and multivariate prediction tasks across a prediction horizon of 4-30 days. 
% Out of both, FEDformer is the current state-of-the-art method for medium to long term time-series prediction. 

% Table generated by Excel2LaTeX from sheet 'Table 4 Paper'
\begin{table*}[htbp]
  \centering
  \caption{Summary of related work on Smart Grid Load Forecasting}
  % \begin{adjustbox}{width=1.8\columnwidth}
    \begin{tabular}{p{8em}|p{6em}|p{7em}|p{5em}|p{6em}|c|c|c} \toprule [.15em]
     \multirow{2}{*}{\textbf{Scheme}} & \multirow{2}{*}{\minitab[l]{\textbf{Training} \\\textbf{Framework} \\\textbf{Used}}} & \multirow{2}{*}{\minitab[l]{\textbf{Deep Learning} \\ \textbf{Architecture} \\ \textbf{Used}}} & \multirow{2}{*}{\minitab[l]{\textbf{Forecast} \\\textbf{Horizon}\\\textbf{(hours)}}} & \multirow{2}{*}{\minitab[l]{\textbf{Models} \\ \textbf{Trained}}} & \multicolumn{3}{c}{\textbf{Privacy Preservation Methodology \& Analysis}}  \\ \cmidrule{6-8}   
     &       &       &       &       & \textbf{Method Applied} &\textbf{Quantitative}$^\dagger$ & \textbf{Qualitative}$^\ddagger$ \\ \midrule[.1em]
    Tian \textit{et al.} \cite{Tian2019Sbc} & SBCTL & S2S RNN & 1     & Global & $\bigstar$   & $\bigstar$ & $\bigstar$  \\ \midrule
    Sehova \textit{et al.} \cite{Sehova2020Dlf} & Central & S2S RNN & 1, 24 & Global & $\bigstar$   & $\bigstar$     & $\bigstar$  \\ \midrule 
    Fekri \textit{et al.} \cite{Fekri2021Dlf} & Central & BNLSTM & 1 - 200 & Global & $\bigstar$   & $\bigstar$     & $\bigstar$  \\ \midrule
    Yazici \textit{et al.} \cite{Yazici2022} & Central & 1D-CNN, LSTM, GRU & 1, 24 & Global & $\bigstar$   & $\bigstar$     & $\bigstar$  \\ \midrule
    {Ryu \textit{et al.} \cite{Ryu2023QM}} &  {Central}   & {MLP-Mixer}  & {24} & {Global} & {$\bigstar$}    & {$\bigstar$}     & {$\bigstar$}  \\ \midrule
    Taik \textit{et al.} \cite{Taik2020Elf} & FL    & LSTM  & 1     & Global & FL    & \xmark     & \xmark  \\ \midrule
    Li \textit{et al.} \cite{Li2020FLB} & FL    & LSTM  & 1     & Global & FL    & \xmark     & \xmark  \\ \midrule
    Liu \textit{et al.} \cite{Liu2021} & FL    & LSTM & 1, 7  & Global & FL + HE & \textbf{-}    & \textbf{-} \\ \midrule
    Fekri \textit{et al.} \cite{Fekri2022Dlf} & FL    & LSTM  & 1, 24 & Global & FL    & \xmark     & \xmark  \\ \midrule
    {Yang \textit{et al.} \cite{Yang2023FL}} & {FL}    & {SecureBoost}  & {1} & {Global} & {FL}    & {\xmark}     & {\xmark}  \\ \midrule
    {Liu \textit{et al.} \cite{Liu2023FedForecast}} &  {FL}   & {RNN + LSTM + GRU}  & {0.5 - 4} & {Global} & {FL}    & {\xmark}     & {\xmark}  \\ \midrule
    {Husnoo \textit{et al.} \cite{Husnoo2023}} &  {FL}   & {RNN + LSTM + GRU + CNN}  & {1} & {Global} & {FL + DP}    & {\xmark}     & {\cmark}  \\ \midrule
    {Liu \textit{et al.} \cite{Liu2023GBRT}} &  {FL}   & {Prob-GBRT}  & {1} & {Global \& Local} & {FL}    & {\xmark}     & {\xmark}  \\ \midrule
    Sakuma \textit{et al.} \cite{Sakuma2022} & SL    & S2S LSTM + GRU + 1D-CNN & 1, 24 & Global \& Local & SL    & \xmark     & \xmark  \\ \midrule
    Proposed & SL    & S2S FEDformer & 96    & Neighbourhood &  SL + DP & \cmark & \cmark \\ \midrule[.15em]
    \multicolumn{8}{l}{{Training framework abbreviations: \textit{SBCTL}: Similarity based Chained Transfer Learning, \textit{FL}: Federated Learning, \textit{SL}: Split Learning.}}\\
    \multicolumn{8}{l}{{Deep learning model abbreviations: \textit{S2S}: Sequence to Sequence, \textit{RNN}: Recurrent Neural Network, \textit{LSTM}: Long Short-Term Memory, }} \\
    \multicolumn{8}{l}{{\quad\textit{GRU}: Gated Recurrent Unit, \textit{BNLSTM}: Bayesian Normalized LSTM, \textit{CNN}: Convolutional Neural Network, }} \\
    \multicolumn{8}{l}{{\quad\textit{MLP}: Multi Layer Perceptron, \textit{Prob-GBRT}: Probabilistic Gradient-Boosted Regression Trees.}} \\
    \multicolumn{8}{l}{\textit{HE}: Homomorphic Encryption, \textit{DP}: Differential Privacy.} \\
    \multicolumn{8}{l}{$\bigstar$: Privacy was not considered while the central model was trained.
    \cmark: Included. \xmark: Not included. \textbf{-} : Not applicable.} \\
    \multicolumn{8}{l}{$\dagger$ Quantitiative: Similarity analysis between clients' data and shared information.} \\
    \multicolumn{8}{l}{$\ddagger$ Qualitative: Effects of privacy preservation approach on model performance.}
    \end{tabular}%
  % \end{adjustbox}
  \label{tab:RelWork}%
\end{table*}%
%%%%%%%%%%%%%%%%%%%%%%%%%%%%%%%%%%%%%%%%%%%%%%%%%%%%%
\subsubsection{Decentralized Learning Methods} \label{sec:DLM}
In their primary forms, both FL and SL frameworks assume that a single model can capture trends across diverse clients; thus, for the load forecasting application, these naive approaches try to learn a single model capable of generating load traces for each client. This, however, is not optimal as the pattern diversity between the clients is usually large, and learning a single forecast model may lead to inferior performance. When used for load forecasting using smart meter data, the methods reviewed so far learn one individual model for each smart meter client or one for a particular group. However, this training strategy becomes computationally expensive as the number of clients grows and raises privacy and security vulnerabilities associated with centralized data transfer.

Several FL and SL-based methods have been proposed to address these issues. In \cite{Taik2020Elf}, an FL-based method has been presented for short-term (One-hour) load forecasting for smart meters with similar load profiles. Their approach uses LSTM as the learning model and federated averaging architecture with weighted averaging for model weights aggregation. The method is shown to work well for short-term (one hour ahead) predictions. Similarly, \cite{Tian2019Sbc} presents a similar approach for short-term forecasting with an emphasis on providing security to the framework via encryption schemes. This, however, leads to increased time complexity of the model. In contrast to \cite{Taik2020Elf}, authors in \cite{Fekri2022Dlf} compare the performance of two FL techniques, FedSGD (single gradient descent step per client) and FedAVG (multiple gradient updates before merging), and allow their clients to have profiles from different distributions. 
Similarly, recent studies like \cite{Liu2023FedForecast} and \cite{Husnoo2023} have leveraged RNN, LSTM, and GRU architectures to train global models for short-term forecasting within the FL framework. Furthermore, in the work by \cite{Husnoo2023}, differential privacy techniques were applied to obscure signs of shared client gradients, providing an additional layer of protection for client privacy.
{Similarly, in \cite{Wang2024R3}, the authors combine FL with the principles of transfer learning to perform demand side forecasting using a Transformer model.}
In {another} work \cite{Liu2023GBRT}, the authors introduce an FL-based boosted multi-task learning framework tailored for inter-district collaborative load forecasting (1 hour ahead). The approach revolves around initially training a central model, which is subsequently employed by individual districts to train personalized models capable of capturing their respective local temporal dynamics. A notable feature of this framework is its use of the probabilistic Gradient-Boosted Regression Tree (GBRT) as the base learner.

{Moving on to works that leverage for privacy preservation, in \cite{Feng2024R1}, authors integrate the FL strategy with local DP to protect user data in recommendation systems. Similarly, in \cite{Guo2025R2}, authors enhance user privacy in task-oriented semantic communication within the 6G landscape by incorporating DP and encryption during the training of a deep neural network-based joint source and channel coding (DeepJSCC) model.}
In \cite{Abuadb2020Cwu}, authors use the SL framework to split a 1D CNN network model into two halves and use it to detect heart abnormalities from the medical ECG dataset. 
Furthermore, they show that in the case of 1D CNN, SL may fail to protect the {patients}' private raw data. To mitigate this data leakage, they use differential privacy (DP) \cite{Dwork2014Taf}, where carefully computed noise is added to the {patients}' activations as an additional layer of security. 
% and test with different number of layers in clients' half of the network. 
Their results show that DP does reduce privacy leakage but at the expense of model performance.
Similar to \cite{Abuadb2020Cwu}, another recent SL-based method \cite{Jiang2022LES} splits an LSTM network to train a classifier for time-series data of multiple patients. To reduce privacy leakage, differential privacy has been used to break the 1-1 relationship between input and its split activations. 

A short summary of related work on the smart grid load forecasting problem is given in Table \ref{tab:RelWork}.
%%%%%%%%%%%%%%%%%%%%%%%%%%%%%%%%%%%%%%%%%%%%%%%%%%%%%%%%%%%
%%%%%%%%%%%%%%%%%%%%%%%%%%%%%%%%%%%%%%%%%%%%%%%%%%%%%%%%%%%
\subsection{Problem Description and Motivation} \label{sec:Moti}
Consider a scenario where an energy provider company distributes power to multiple neighbourhoods/districts of a city. Each community is served by a single Grid Station (GS). The Service Provider (SP) is interested in training a load forecasting model for medium (few hours) to long-term (few days - weeks) forecasting to better manage their generation capacity and reduce energy waste. To do so, they can employ different strategies, e.g., the SP might want to learn a single prediction model for all districts, which is {easier} but not optimal as households {across neighbourhoods may} have different load profile distributions. Thus, training a single model to cover all distributions {may} lead to significant prediction errors. Conversely, training a single model for every client is cumbersome and infeasible if the number of clients is substantial. Instead of either of these extremes, we propose to learn a single model for each neighbourhood as one would expect clients from the same neighbourhood to have similar load profiles, facilitating the training of an accurate prediction model. 

Next, we need to decide on a training framework, i.e., central or {decentralized}. 
As data privacy is our top priority, {decentralized} learning strategies like FL and SL should be employed where the private data never leaves the client's premises. However, as the client-side training has to be performed by a smart meter, the training process's computational and data transfer requirements have to be modest so as not to hinder their main functionalities.
The SL framework is selected to ensure this due to its low computational and communications requirements and privacy-preserving nature. 
%%%%%%%%%%%%%%%%%%%%%%%%%%%%%%%%%%%%%%%%%%%%%%%%%%%%%%%%%%%
\subsection{Contributions}
The major contributions of this paper are as follows:
\begin{itemize}
    \item We propose a novel SL framework with a dual split strategy, i.e., the network's first split (Split-1) resides at the GS covering a single neighbourhood, and the second split (Split-2) {stays} at the SPs' end. To reduce computational load on smart meters, each client is only responsible for performing a forward pass on its private data using their GSs' Split-1 network, computing the loss, and initiating back-propagation. GS is responsible for carrying out back-propagation through the Split-1 model and updating its weights. 
    The models, as a whole, are trained using two alternative strategies: \textit{SplitGlobal}, which trains unique Split-1 models for each neighbourhood and a global Split-2 model shared by all neighbourhoods, and \textit{SplitPersonal}, which trains personalized split models (Split-1 and Split-2) for each neighbourhood. Once the training is complete, the SP will have access to both network splits and can perform individual-level (requiring respective clients' involvement) and neighbourhood-level predictions using the cumulative load trend from the respective neighbourhood GS.
    \item As our base learner, we utilize {the} transformer \cite{Vaswan2017Aia} based architecture, called FEDformer \cite{Zhou2022FED}. Compared with widely used LSTM, transformers enjoy better performance and can fully utilize the acceleration offered by discrete graphics processing units.
    Based on our literature review, ours is the first work that uses a transformer architecture to implement split learning for electricity load forecasting. Extensive experiments are conducted to assess the performance of the trained split model against a centrally trained model under multiple scenarios.   
    \item We present a detailed quantitative assessment of the extent of information leakage between clients' private data and their respective Split-1 activations using mutual information-based neural estimation (MINE) \cite{Belgha2018Mmi}. Based on the estimated mutual information (MI) between input and activations, a vigilant client can decide whether the current activations batch is secure enough to be forwarded to the GS or not. For additional privacy, we incorporate {differential privacy \cite{Dwork2014Taf, Abadi2016Dlw}} to further obfuscate the client's Split-1 activations into the model framework and analyze its effects on information leakage and model performance.
\end{itemize}
The rest of the paper is organized as follows:
Section \ref{sec:Prelim} briefly describes FEDformer,
the transformer variant used in this work. We further discuss the concepts of split learning, differential privacy, and mutual information neural estimation. 
Section \ref{sec:Proposed} outlines the proposed system model, the FEDformer model split and the SL training framework. Section \ref{sec:ExpEval} presents experimental evaluations of the proposed framework under different testing scenarios, followed by privacy leakage analysis using mutual information and differential privacy.
We conclude the paper in Section \ref{sec:Conc}.
%================================================================================================
%================================================================================================
%%%%%%%%%%%%%%%%%%%%%%%%%%%%%%%%%%%%%
\section{Preliminaries} \label{sec:Prelim}
In this section, we briefly describe 
% the original transformer architecture \cite{Vaswan2017Aia}, 
the frequency enhanced attention blocks proposed in FEDformer \cite{Zhou2022FED}, split learning framework \cite{Vepako2018Slf}, differential privacy for machine learning \cite{Dwork2014Taf}, and mutual information neural estimation \cite{Belgha2018Mmi} for privacy leakage analysis.
%%%%%%%%%%%%%%%%%%%%%%%%%%%%%%%%%%%%%%%% Vanilla transformer section removed due to size constraints
% Moved to Appendix file
\subsection{FEDformer} \label{sec:FEDformer}
FEDformer follows the deep decomposition architecture proposed in Autoformer \cite{Wu2021Autoformer}, where the input time series is analyzed by decomposing it into a seasonal and a trend-cyclical part. The seasonal part is expected to capture seasonality, whereas the trend-cyclical part is expected to capture the long-term temporal progression of the input. FEDformer uses a series decomposition block with a single or a set of moving average filters (of different sizes) to perform such decomposition. 
% The FEDformer architecture (in split form) is given in Fig. \ref{fig:Split_Model}. 
FEDformer implements self-attention mechanisms in the frequency domain using two distinct blocks, a Frequency Enhanced Block (FEB) and a Frequency Enhanced Attention (FEA) Block. Their working is briefly discussed next.
%%%%%%%%%%%%%%%%%%%%%%%%%%%%%%%%%%%%%%%%%%%%%%%%%%%
\subsubsection{Frequency Enhanced Block} \label{sec:FEB}
In \cite{Zhou2022FED}, the authors proposed Fourier transform and Wavelet transform to work in the frequency domain. Here we will only focus on the Fourier transform. Let $\textbf{X}\in\mathbb{R}^{L\times D}$ be the input to the FEB, where $L$ is the sample length, and $D$ is the inner dimension of the model. Query matrix \textbf{Q} is computed by linearly projecting \textbf{X} with $\textbf{W}\in\mathbb{R}^{D\times D}$ as $\textbf{Q}=\textbf{X}\,\textbf{W}$. Next, \textbf{Q} is transformed from time to frequency domain using Discrete Fourier transform (DFT) to get $\mathcal{Q}\in\mathbb{C}^{L\times D}$. A subset of randomly selected Fourier components is discarded from $\mathcal{Q}$ to get a reduced dimensional matrix $\tilde{\mathcal{Q}}\in\mathbb{C}^{M\times D}$, where $M < L/2$. Finally, the output of FEB is computed as
\begin{equation}
    \textrm{FEB}(\textbf{X}) = \mathcal{F}^{-1}(\textrm{ZeroPad}(\tilde{\mathcal{Q}} \bigodot \textbf{R})
\end{equation}
where $\textbf{R}\in\mathbb{C}^{D\times D\times M}$ is a randomly initialized parametric kernel, and $\bigodot$ is the production operator. The result of $(\tilde{\mathcal{Q}} \bigodot \textbf{R}) \in \mathbb{C}^{M\times D}$ is then zero-padded to $\mathbb{C}^{L\times D}$ and transformed back into time domain via inverse DFT.
%%%%%%%%%%%%%%%%%%%%%%%%%%%%%%%%%%%%%%%%%%%%%%%%%%%
\subsubsection{Frequency Enhanced Attention Block} \label{sec:FEA}
The FEA block takes two inputs, the encoder output $\textbf{X}_{en}$ and $\textbf{X}_{de}$ from decoder, and generates the query \textbf{Q} matrix via linearly projecting $\textbf{X}_{de}$ using weight matrix $\textbf{W}_q$, whereas the key \textbf{K}, and value \textbf{V} matrices are generated by projecting $\textbf{X}_{en}$ using $\textbf{W}_k$ and $\textbf{W}_v$, respectively. 

The query, key, and value matrices are then transformed from time to frequency domain via DFT followed by random mode selection (as in FEB Section \ref{sec:FEB}) to get $\tilde{\mathcal{Q}}, \tilde{\mathcal{K}}, \tilde{\mathcal{V}} \in\mathbb{C}^{M\times D}$. Finally, the output of FEA is computed as
\begin{equation} \label{eq:FEA}
    \textrm{FEA}(\textbf{Q}, \textbf{K}, \textbf{V}) = \mathcal{F}^{-1}(\textrm{ZeroPad}(\sigma(\tilde{\mathcal{Q}}\, \tilde{\mathcal{K}}^\top) \tilde{\mathcal{V}})),
\end{equation}
where $\sigma$ is the \textit{tanh} activation function.
%%%%%%%%%%%%%%%%%%%%%%%%%%%%%%%%%%%%%%%%%%%%%%%%%%%
\subsection{Split Learning} \label{sec:SplitLearning}
In split learning (SL) \cite{Vepako2018Slf}, a deep neural network (DNN) is split into two halves; clients maintain the first half and the remaining layers are maintained by a server. Consequently, a group of clients are able to train a DNN collaboratively using (but not sharing) their collective data. Additionally, the server performs most of the computational work, reducing the clients' computational requirements. However, this comes at the cost of privacy trade-off, i.e., the output of earlier layers leaks more information about the inputs \cite{Erdoga2021SDa}. 
Here, choosing a suitable split size is important for expecting data privacy as it has been shown that for a relatively small client model, an {honest-but-curious \cite{Paverd2014HBCA}} server can extract {the clients' private} data accurately just by knowing the client-side model architecture \cite{Abuadb2020Cwu}. Thus, it is recommended that in SL, clients should compute more layers, increasing computational load but incurring stronger privacy \cite{Abuadb2020Cwu}.

During training, clients perform forward passes using their own data up to the final split layer of DNN. These activations are then shared with the server, which continues the forward pass on its DNN split. If the label sharing between clients and servers is enabled, the server can compute the loss itself. Otherwise, it has to send the activations of its final layer back to the client for loss computation. In this case, the gradient backpropagation begins on the client side, and the client feeds the gradient back to the server, which continues backpropagation through its DNN split. Finally, the server shares the gradients at its first layer with the client, who finishes the backpropagation through its DNN split. When more than one client is participating in training, SL adds all clients into a circular queue, whereby each client takes turns using their private data to train with the server. At the end of a training round, the client shares its updated model weights with other clients either through a central server, directly with each other, or via a P2P network.
%%%%%%%%%%%%%%%%%%%%%%%%%%%%%%%%%%%%%%%%%%%%%%%%%%%
\subsection{Differential Privacy}  \label{sec:DiffPriv}
{One of the most widely used privacy-preserving technologies is Differential Privacy (DP) \cite{Dwork2014Taf}. Its effectiveness in safeguarding user data privacy has been extensively demonstrated by adding carefully computed noise to the data.
The machine learning community has widely used DP to ensure data privacy \cite{Ji2014, Abadi2016Dlw}. Let $\mathcal{X}$ be the input space, $\mathcal{Y}$ the output space, $\epsilon$ the privacy budget parameter, $\delta \in o(\frac{1}{n})$ be a non-negative heuristic parameter, $n$ be the number of samples in the dataset, and $\mathcal{M}$ a randomization mechanism. We say that the mechanism $\mathcal{M}: \mathcal{X} \rightarrow \mathcal{Y}$ is $(\epsilon,\delta)-$differentially private ($(\epsilon,\delta)$-DP) if, for any neighbouring datasets $\textbf{X}_1$ and $\textbf{X}_2$ (differing by a single element) in $\mathcal{X}$, and any output $S\subseteq \mathcal{Y}$, as long as the following probabilities are well-defined, there holds
\begin{equation}
    Pr[\mathcal{M}(\textbf{X}_1)\in S] \le e^\epsilon \times Pr[\mathcal{M}(\textbf{X}_2)\in S ] + \delta.
    \label{eq:DP_1}
\end{equation}
Intuitively, \eqref{eq:DP_1} provides an upper bound ($e^\epsilon$) on the difference between outputs of the mechanism $\mathcal{M}$ when applied to two neighbouring datasets, where the value of $\epsilon$ controls the overall strength of the privacy mechanism and $\delta$ accounts for the probability that privacy might be violated \cite{Abadi2016Dlw}. Thus, to ensure stronger privacy protection, both $\epsilon$ and $\delta$ should be kept low. With $\delta=0$, the pure $\epsilon-$DP is shown to be much stronger than the $(\epsilon,\delta)-$DP (with $\delta>0$) in terms of mutual information \cite{De2012epDP}. 
{The addition of $\delta$ in the formulation is to provide a level of plausible deniability, allowing for a small probability ($\delta$) that an individual's data might be exposed or identified by an attacker. While $\epsilon$ governs the average privacy loss incurred, $\delta$ plays a role in controlling the worst-case privacy loss scenario \cite{Ji2014}.
Additionally,} the $(\epsilon, \delta)-$DP offers the advantage of advanced composition theorems, enabling a substantially greater number of training iterations compared to pure $\epsilon-$DP with the same $\epsilon$. As a result, most recent works in differentially private machine learning have shifted away from $\epsilon-$DP.}

Let $f:\mathcal{X}\rightarrow \mathbb{R}$ be a deterministic real-valued function. Then, in order to approximate this function with an {$(\epsilon,\delta)-$DP} mechanism, noise calibrated with \textit{f}'s sensitivity ($s$) is added to its output. Here, sensitivity is defined as $s=\max_{\textbf{X}_1, \textbf{X}_2}\Vert f(\textbf{X}_1) - f(\textbf{X}_2) \Vert$. Intuitively, for high sensitivity, it is much easier for an adversary to extract information about the input \cite{Dwork2014Taf}. 
The general mechanism that satisfies the DP is defined by
\begin{equation} \label{eq:LDP}
    \mathcal{M}(\textbf{X}) \triangleq f(\textbf{X}) + \eta
\end{equation}
where $\eta$ is a random variable from distribution $\mathcal{N}(0, \frac{2s^2}{\epsilon^2}\log(\frac{2}{\delta}))$ under $(\epsilon,\delta)-$DP or $p(\eta) \propto e^{-\epsilon\Vert\eta\Vert/s}$ under $\epsilon-$DP \cite{Ji2014}.
In our case, $f$ is the split model, $\textbf{X}$ is a clients' private input dataset, sensitivity $s$ is computed across the batch axis of the Split-1 activations tensor, and the noise is added to the clients' batch activations at Split-1 model output.
In this way, the noise level is controlled by {the sensitivity} $s$, which comes from the data, the probability $\delta$, and the privacy budget $\epsilon$, which can be set according to the privacy requirements, e.g., $\epsilon=10$ results in a weak privacy guarantee as compared to $\epsilon \approx 0$, which gives the strongest privacy guarantee but makes the data useless. 
The privacy guarantee of a DP mechanism \eqref{eq:LDP} is that the likelihood of revealing sensitive information about any individual in the input dataset through the algorithm's output is significantly reduced \cite{Ji2014}.
In our study, we analyze both DP mechanisms in terms of the mutual information leakage between input and output of the clients' split network.

%%%%%%%%%%%%%%%%%%%%%%%%%%%%%%%%%%%%%%%%%%%%%%%%%%%
\subsection{Mutual Information Neural Estimation} \label{sec:MINE}
{The clients' split layer activations in SL frameworks are shared with the server.} However, these activations may carry enough information about the input that an adversary might be able to precisely reconstruct the original data \cite{Pasqui2021Utt}. Various researchers have utilized noise addition mechanisms offered by {$(\epsilon, \delta)$-DP} as a security measure to mitigate this issue. However, it is difficult to quantify the relationship between the added noise level and the information leakage risk. 
Mutual information (MI) is commonly used in information theory to assess how much information can be inferred from one random variable (RV) about another. Compared to correlations, MI can capture non-linear statistical dependencies between RVs \cite{Belgha2018Mmi}; however, it is difficult to compute, especially for high-dimensional RVs. In \cite{Belgha2018Mmi}, the authors have proposed to compute MI using neural networks using the fact that MI between two IID RVs \textbf{X} and \textbf{Y} is equivalent to the Kullback-Leibler Divergence (KLD) between their joint ($\mathbb{P}_{\textbf{XY}}$) and product of their marginal ($\mathbb{P}_\textbf{X} \otimes \mathbb{P}_\textbf{Y}$) distributions. According to the Donsker-Varadhan representation of KLD \cite{Donske1983Aeo}, the MI between \textbf{X} and \textbf{Y} is lower bound by
\begin{eqnarray} \label{eq:MINE}
    I(\textbf{X};\textbf{Y}) &=& D_{KL}\left[ \mathbb{P}_{\textbf{XY}} || \mathbb{P}_\textbf{X} \otimes \mathbb{P}_\textbf{Y} \right] \nonumber \\
    &\ge& \sup_{T\in \mathcal{T}} \mathbb{E}_{\mathbb{P}_{\textbf{XY}}}[T] - \log(\mathbb{E}_{\mathbb{P}_\textbf{X} \otimes \mathbb{P}_\textbf{Y}}[e^T]).
\end{eqnarray}
Here $\mathcal{T}$ is any class of functions $T:(\textbf{x}_i, \textbf{y}_i) \rightarrow \mathbb{R}$ that satisfies the integrability constraints of Donsker-Varadhan theorem. 
Under this setting, the authors in \cite{Belgha2018Mmi} use a neural network to model $\mathcal{T}$, which converts the MI problem to a network optimization one, leveraging neural networks' ability to approximate arbitrary complex functions. 

Consider an SL framework where $\textbf{x}$ and $\textbf{y}$ are the batched inputs and outputs of the split network, and we are interested in finding how much information about \textbf{x} can be inferred from \textbf{y}. To do so, we need to estimate the MI between them. Ref. \cite{Belgha2018Mmi} says that this MI is lower bounded by \eqref{eq:MINE}. Let $T$ be a neural network. Then, the expectations in \eqref{eq:MINE} are empirically estimated by sampling from joint distribution as $(\textbf{x},\textbf{y}) \sim \mathbb{P}_\textbf{XY}$ and from marginals by shuffling \textbf{y} across the batch axis to get $(\textbf{x},\overline{\textbf{y}})$. In other words, in $(\textbf{x},\textbf{y})$ the input-output relationship is intact, whereas, in $(\textbf{x},\overline{\textbf{y}})$, this relationship has been broken. The network $T$ is trained by maximizing \eqref{eq:MINE}. Thus, if MI between a batched \textbf{x} and \textbf{y} is large, \eqref{eq:MINE} computed using a trained network $T$ will be high, and vice versa. We demonstrate this effect in detail in Section \ref{sec:InfoLeak}.
% The network architecture can be designed to train on any type of data using samples from the joint and marginal distributions of respective RVs. 
%%%%%%%%%%%%%%%%%%%%%%%%%%%%%%%%%%%%%%%%%%%%%%%%%%%
%%%%%%%%%%%%%%%%%%%%%%%%%%%%%%%%%%%%%%%%%%%%%%%%%%%
%================================================================================================
%================================================================================================
\begin{figure}
    \centering
    \includegraphics[width=\columnwidth]{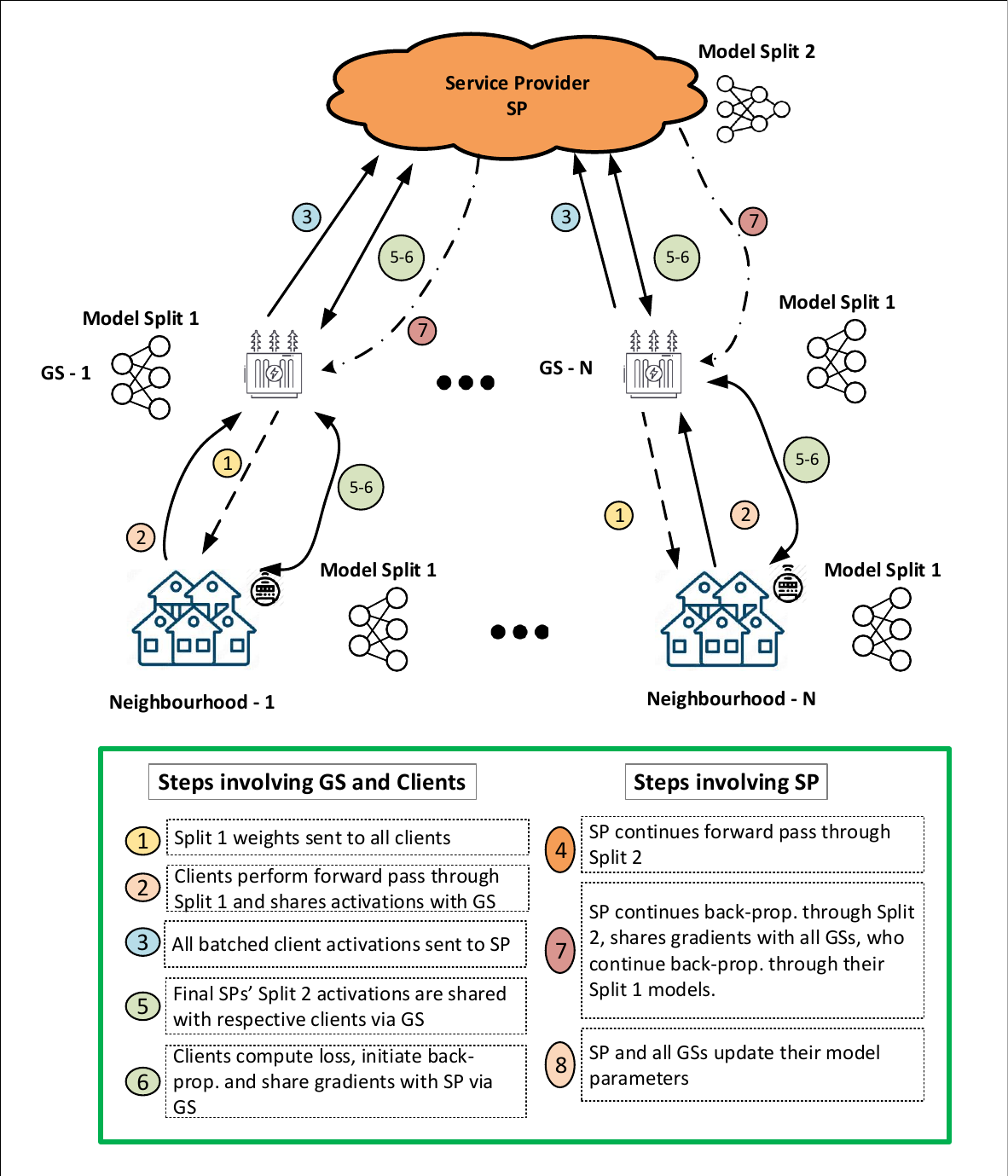}
    %\caption{The proposed split learning framework. The electricity provider is represented as the Service Provider, which can directly communicate with all Grid Stations. These Grid Stations distribute power to a specific neighbourhood where each client is equipped with a smart meter. Grid Stations can also communicate directly with each smart meter. The number next to each arrow represents the individual steps of one training-batch epoch as described in Section \ref{sec:PropFW}.}
    \caption{{The proposed system model and split learning framework. SP is the service provider, and GS is the grid station.
    }}
    \label{fig:SL_Train}
\end{figure}
%================================================================================================
\section{Proposed Split Learning Framework for Electricity Load Forecasting} \label{sec:Proposed}
In this section, we first discuss the proposed system model, the adversarial model, the FEDformer model split and their internal modules, followed by our proposed split learning framework and its training methodology.
%%%%%%%%%%%%%%%%%%%%%%%%%%%%%%%%%%%%%%%%%%%%%%%%%%%
\subsection{System Model} \label{sec:SysModel}
We consider a three-tier system model with four major entities, as shown in Fig. \ref{fig:SL_Train}. On top, we have the electricity Service Provider; in the middle, we have Grid Stations responsible for distributing electricity to the individual districts/neighbourhoods. The lowest tier comprises smart meter clients (industrial, commercial, or residential) lumped into neighbourhoods. 
Under this model, the SP is an organization which procures electricity from various providers and is responsible for meeting the energy requirements of all connected GSs.
Moreover, the smart meters can connect to their GS using a secure communications protocol, e.g., cellular network or power-line communications. The GSs are connected to SP via a private network or the Internet. The SP can not connect directly with any client and has to go through the GS to communicate with a client.
The objective is to learn a DL time series prediction model, using a split learning framework, on all clients' data without compromising the individual clients' privacy. 
Once trained, the entire model will be accessible to the SP, and the SP can effectively perform medium to long-term load forecasting for any client from any neighbourhood.
%%%%%%%%%%%%%%%%%%%%%%%%%%%%%%%%%%%%%%%%%%%%%%%%%%%
\subsection{Adversary Model} \label{sec:AdvModel}
In this paper, we assume a modest security environment, i.e., the GS, SP, and the clients are honest-but-curious {\cite{Paverd2014HBCA, Abuadb2020Cwu}}. Thus, the participants may not try to poison the training process; however, both the GS and SP may collude to infer information about clients' private data as they have access to the entire model and the client-side split layer activations (details in Section \ref{sec:PropFW}). Furthermore, external adversaries or some malicious clients may try to intercept the split layer activations sent by the other clients to GS to extract their private data. Thus, the attack model considered here is the model inversion attack, which aims to extract clients' sensitive data given only their activations. Under this attack, the objective of the adversary is to find a function $\mathcal{G}$ which can infer the client's private data $\textbf{X}$ from its split activations $\textbf{A}$ as $\textbf{X} = \mathcal{G} (\textbf{A})$. However, in practice, this inference need not be exact, as a close approximation of the client's data is usually enough.
%%%%%%%%%%%%%%%%%%%%%%%%%%%%%%%%%%%%%%%%%%%%%%%%%%%
\begin{figure*}
    \centering
    \includegraphics[width=2\columnwidth]{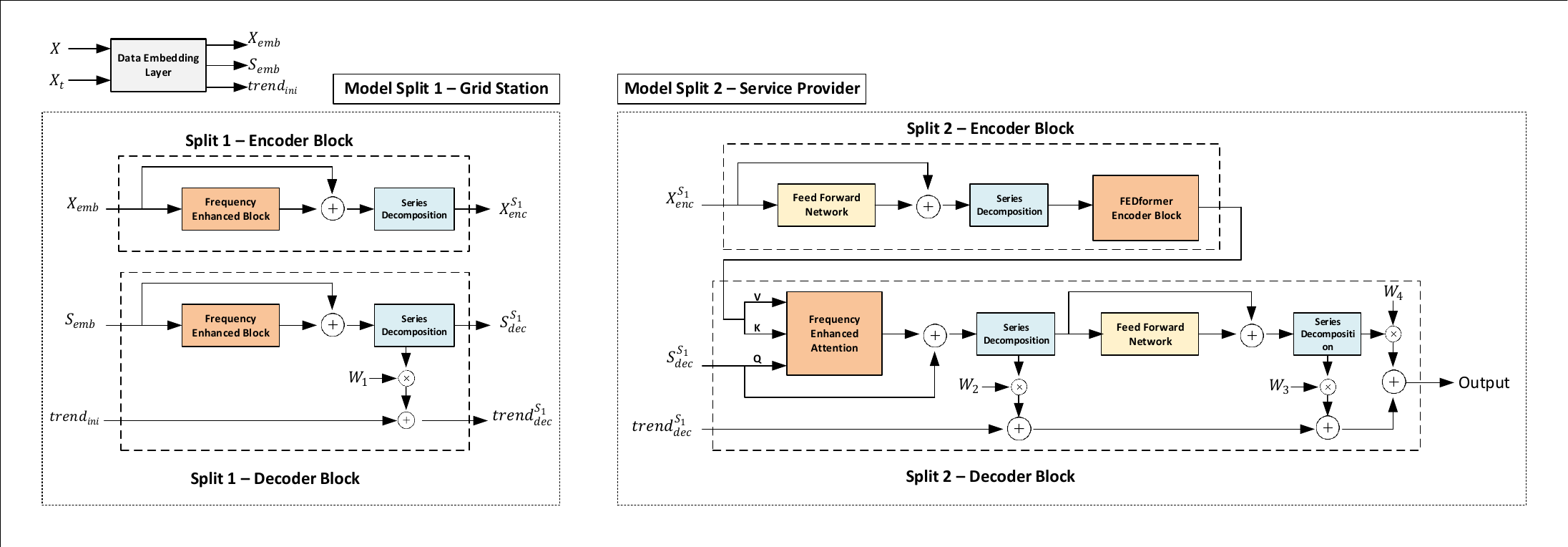}
    \caption{Dual split FEDformer model.}
    \label{fig:Split_Model}
\end{figure*}
\subsection{The FEDformer Model Split} \label{sec:ModSplit}
We have selected FEDformer architecture \cite{Zhou2022FED} (with some modifications) as our backbone prediction model, where the entire model consists of two encoders and a single decoder block. The FEDformer model is split into two halves, termed Split-1 and Split-2. The Split-1 model contains the first two inner blocks of the FEDformer encoder and FEDformer decoder, whereas the Split-2 model contains the remaining inner blocks of the first FEDformer encoder followed by a complete FEDformer encoder block and remaining inner blocks of the FEDformer decoder.
Each GS gets its own copy of the Split-1 model, and the SP maintains Split-2. The resulting split FEDformer is shown in Fig. \ref{fig:Split_Model}. In Section \ref{sec:FEDformer} we have discussed the series decomposition, FEB and FEA blocks, and the rest are briefly discussed next.

The \textit{data embedding layer} included in model Split-1, as shown in Fig. \ref{fig:Split_Model}, consists of a series decomposition layer, a 1D convolutional layer, and two linear layers. It takes three inputs, an input time-series of length L $\textbf{X}\in\mathbb{R}^{L\times Z}$, where $Z$ is the dimension of each time point (for univariate case, $Z=1$), the timestamp encoded information of the input time-series as $\textbf{X}_t\in\mathbb{R}^{L\times U}$, and output time-series $\textbf{Y}_t\in\mathbb{R}^{(L/2+O)\times U}$, where $U$ depends upon the temporal granularity, and $O$ denotes the prediction time horizon, for instance, for the scale of Year-Month-Day-Hour, $U=4$. 
First, $\textbf{X}$ is passed through the \textit{series decomposition block} to get a seasonal component $\textbf{X}_S$ and a trend component $\textbf{X}_{trend}$. Then, the embedded inputs to the encoder and decoder blocks of Split-1 are generated as:
\begin{eqnarray}
    \textbf{X}_{emb} \!\!\!&=&\!\!\! 1DConv(\textbf{X}) + Linear(\textbf{X}_t) \nonumber \\
    \textbf{S}_{emb} \!\!\!&=&\!\!\! 1DConv(Concat(\textbf{X}_{S,L/2:L}, \textbf{X}_{0})) + Linear(\textbf{Y}_t) \nonumber\\
    trend_{ini} \!\!\!&=&\!\!\! Concat(\textbf{X}_{trend,L/2:L}, \textbf{X}_{mean}),
\end{eqnarray}
where $\textbf{X}_0, \textbf{X}_{mean}\in\mathbb{R}^{O\times Z}$ denote the placeholders filled with zeros and mean of \textbf{X}, respectively, $\textbf{X}_{emb}\in\mathbb{R}^{L\times D}$, $\textbf{S}_{emb} \in \mathbb{R}^{(L/2+O) \times D}$, and $trend_{ini}\in\mathbb{R}^{(L/2+O)\times Z}$. Here, $D$ is the inner dimension of the model. In this way, the decoder takes guidance from the later half of the input time series to fill in the remaining placeholder data points during training. 
Since the series-wise connection will inherently keep the sequential information, we do not need to perform position embedding, which differs from vanilla Transformers.
This specific setting for the Split-1 network was chosen to keep the computational requirements low while ensuring a complex non-linear relationship between the input and outputs of Split-1 to ensure low information leakage between them (see details in Section \ref{sec:InfoLeak}).

The feed-forward network, seen in the Split-2 encoder and decoder, is a two-layer fully connected neural network (FCNN) with input and output dimensions $D$, and inner dimension $D_{ff}$. The output of its first layer is passed through \textit{GELU} activation function before passing to layer two. 
The output (seasonal and trend) tensors generated by series decomposition blocks found in Split-1 and Split-2 decoders have dimensions $D$, whereas the incoming $trend_{ini}$ and $trend_{dec}^{S_1}$ containing temporal trend information are $Z$ dimensional (dimension of the input time series). Thus, the trend output tensor of a series decomposition block is first projected back into $Z$ dimensional tensors using trainable projection matrices $\textbf{W}_i\in\mathbb{R}^{D\times Z}$, where $i=\left[1,2,3\right]$, before adding to the incoming trend tensors. Similarly, the seasonal tensor of the final series decomposition block of the Split-2 decoder is also projected down to $Z$ before adding to the incoming trend tensor to generate the final prediction output.
%%%%%%%%%%%%%%%%%%%%%%%%%%%%%%%%%%%%%%%%%%%%%%%%%%
\subsection{Proposed Split Learning Framework} \label{sec:PropFW}
Figure \ref{fig:SL_Train} depicts the SL training process for load forecasting. Note that the clients do not transfer prediction targets (labels) to the GS or SP. As discussed in Section \ref{sec:ModSplit}, the FEDformer is split into two halves, and each GS maintains a personalized copy of model Split-1. The model Split-2 resides at the SP. 
Unlike a traditional Split Learning framework where clients are responsible for the forward pass, backpropagation through their network split, and parameter updates, our framework proposes a different approach. In our proposed framework, smart meters perform only the forward pass through their data, compute loss (after receiving predictions from SP through GS), and initiate backpropagation (gradient computation of loss with respect to the predictions only). {The GS performs backpropagation through Split-1 and parameter updates, significantly reducing the computational load on the smart meters.}
As a byproduct, we do not need to over-simplify the Split-1 model; thus, non-linear relationships between clients' inputs and activations are maintained, ensuring stronger privacy.

Two strategies are employed when training at SP, i.e., SP can learn a single Split-2 model for all GSs (neighbourhoods) or one personalized Split-2 model per GS. In the case of the former, the overall learned model is referred to as \textit{SplitGlobal}, whereas the latter model is called \textit{SplitPersonal}. The goal is to analyze the generalization capabilities of the model when performing predictions for clients from the same as well as those coming from different neighbourhoods, as clients from different neighbourhoods are expected to have different load patterns and distributions. Nevertheless, the training process is relatively similar for both models. Algorithm \ref{algo:A1} outlines the pseudo-code for training the \textit{SplitGlobal} variant of the proposed SL model.

We start with weight initialization for all SP and GS models. At the start of each training epoch, A GS selects all or a subset $C_t$ of $K$ neighbourhood clients (chosen randomly or the same as the previous epoch).
% SP initialises its own Split-2 model(s) and asks each GS to initialize their individual Split-1 models as well. 
Following this, the training begins in parallel across all GSs, where the participating clients use a single private training batch to perform the model update in 7 distinct sequential steps (see Fig. \ref{fig:SL_Train}). These steps are summarized in relation to Algorithm \ref{algo:A1} next. \\
\textbf{Steps 1-2:} Each GS sends its Split-1 model weights to all selected neighbourhood clients $C_t$. In parallel, each receiving client $k$ performs a forward pass through the received Split-1 model using one of its private training batches $b$. The activations of the final split layer, denoted by $\textbf{A}^{GS}_{k,b}$ of Split-1 model are forwarded to the GS (lines 8-11 of Algorithm \ref{algo:A1}). \\
\textbf{Step 3:} Once a GS has received batched activations from all of its training clients, it concatenates their activations, denoted by $\textbf{A}_b^{GS}$, and forwards them to the SP to continue the forward pass (lines 12-13 of Algorithm \ref{algo:A1}). \\
\textbf{Step 4:} Depending upon the model being trained, i.e., \textit{SplitGlobal} or \textit{SplitPersonal}, this step is performed little differently. In the case of \textit{SplitGlobal}, the SP performs the forward pass through a single Split-2 model using a batch size of $K\times (clients\, batch\, size)$, where $K$ denotes the number of clients (lines 14-15 of Algorithm \ref{algo:A1}). For \textit{SplitPersonal}, the SP performs a forward pass through the individual personalized Split-2 models using the activations received from their respective GSs.  \\
\textbf{Step 5:} As the training data is never allowed to leave the client's premises, the final Split-2 layer outputs are forwarded to their respective clients through GSs for loss computation (lines 17-19 of Algorithm \ref{algo:A1}). \\
\textbf{Step 6:} Each client computes loss, initiates back-propagation and shares gradients w.r.t. outputs $\textbf{O}_{k,b}^{GS}$ with their GS, which forwards these gradients to SP (lines 20-22 of Algorithm \ref{algo:A1}). \\
\textbf{Step 7:} Once gradients from all GSs are received,
SP continues the gradient back-propagation through its Split-2 model(s). The gradients w.r.t. each $\textbf{A}_b^{GS}$ are then forwarded back to their respective GSs. Each GS averages the received gradients across clients' dimensions and continues back-propagating through their respective Split-1 models (lines 24-25 of Algorithm \ref{algo:A1}). \\
\textbf{Step 8:} Once all gradients have been populated, SP and each GS update their model weights (line 26 of Algorithm \ref{algo:A1}).

Following the model updates, the GSs pass their updated models back to their respective clients for another round of batch training. This is repeated until all batches have been iterated over, thus completing a single training epoch. For the next epoch, GSs can continue training with the same clients or select new ones. Once training is finished, each GS will have a personalized Split-1 model.

\begin{algorithm}
    \KwIn{Number of GSs: $nGS$, Client Input: \textbf{X}, Client Output: \textbf{Y}, loss function $\mathcal{L}$.}
    \textbf{Initialization:} 
    Initialize model weights for \textit{Split-1} and \textit{Split-2} $\forall$ GS and SP models. \\
    \For{t in epochs}{
        Each GS selects $C_t \leftarrow$ set of $K$ clients.\\
        Let $nB=\min(nB_1, nB_2, \ldots, nB_K)$, where $nB_k$ is the number of batches available at client k.\\
        \For{b in nB}{
            \textbf{Forward Pass:} \\
            \For{GS in nGS \textbf{in parallel}}{
                GS shares \textit{Split-1} model weights with each client in $C_t$, \\
                \For{k in $C_t$ \textbf{in parallel}}{
                    $\textbf{A}^{GS}_{k,b} \leftarrow Split\textrm{-}1(\textbf{X}_{k,b}^{GS})$, \\
                    $\textbf{A}^{GS}_{k,b} \textrm{is shared with respective GS.}$
                }
                $\textbf{A}^{GS}_{b} = Concat(\textbf{A}^{GS}_{1,b}, \ldots, \textbf{A}^{GS}_{K,b})$ \\
                $\textbf{A}^{GS}_{b}$ is passed to SP.
            }
            For each GS, SP generates and shares:\\ 
            $\quad\textbf{O}_b^{GS}\leftarrow Split\textrm{-}2(\textbf{A}^{GS}_{b})$.\\
            \textbf{Loss Computation:}\\
            \For{GS in nGS \textbf{in parallel}}{                
                \For{k in $C_t$ \textbf{in parallel}}{
                    From $\textbf{O}_b^{GS}$, GS sends respective outputs batch $\textbf{O}_{k,b}^{GS}$ to each client $k$, \\
                    $\ell_{k,b}^{GS} = \mathcal{L}(\textbf{Y}_{k,b}^{GS} - \textbf{O}_{k,b}^{GS}),$ \\
                    Each client initiates Back-prop. and shares gradients with their GS.
                }                
                GS shares the received gradients with SP.\\
            }
            \textbf{Back-prop. \& Model Update:}\\
            SP continues back-prop. through \textit{Split-2} and shares gradients of first \textit{Split-2} layer w.r.t. $\textbf{A}^{GS}_{b}$ with respective GSs. \\
            Each GS resumes back-prop. through their \textit{Split-1} model. \\
            SP and each GS perform model update.
        }
    }
    \KwOut{Each GS gets a trained \textit{Split-1} model and SP gets trained \textit{Split-2} model.}
    \caption{Pseudocode for \textit{SplitGlobal} model training under the proposed SL framework.}
    \label{algo:A1}
\end{algorithm}
%%%%%%%%%%%%%%%%%%%%%%%%%%%%%%%%%%%%%%%%%%%%%%%%%%%
%%%%%%%%%%%%%%%%%%%%%%%%%%%%%%%%%%%%%%%%%%%%%%%%%%%
%================================================================================================
%================================================================================================
\section{Experimental Evaluation} \label{sec:ExpEval}
In this section, we provide a detailed experimental evaluation of the proposed framework under multiple testing scenarios and present the mutual information-based privacy leakage analysis resulting from sharing activations with and without differential privacy. 
%%%%%%%%%%%%%%%%%%%%%%%%%%%%%%%%%%%
\subsubsection{Dataset and Evaluation Metrics} \label{sec:Dataset}
\begin{figure}
    \centering
    \includegraphics[width=\columnwidth]{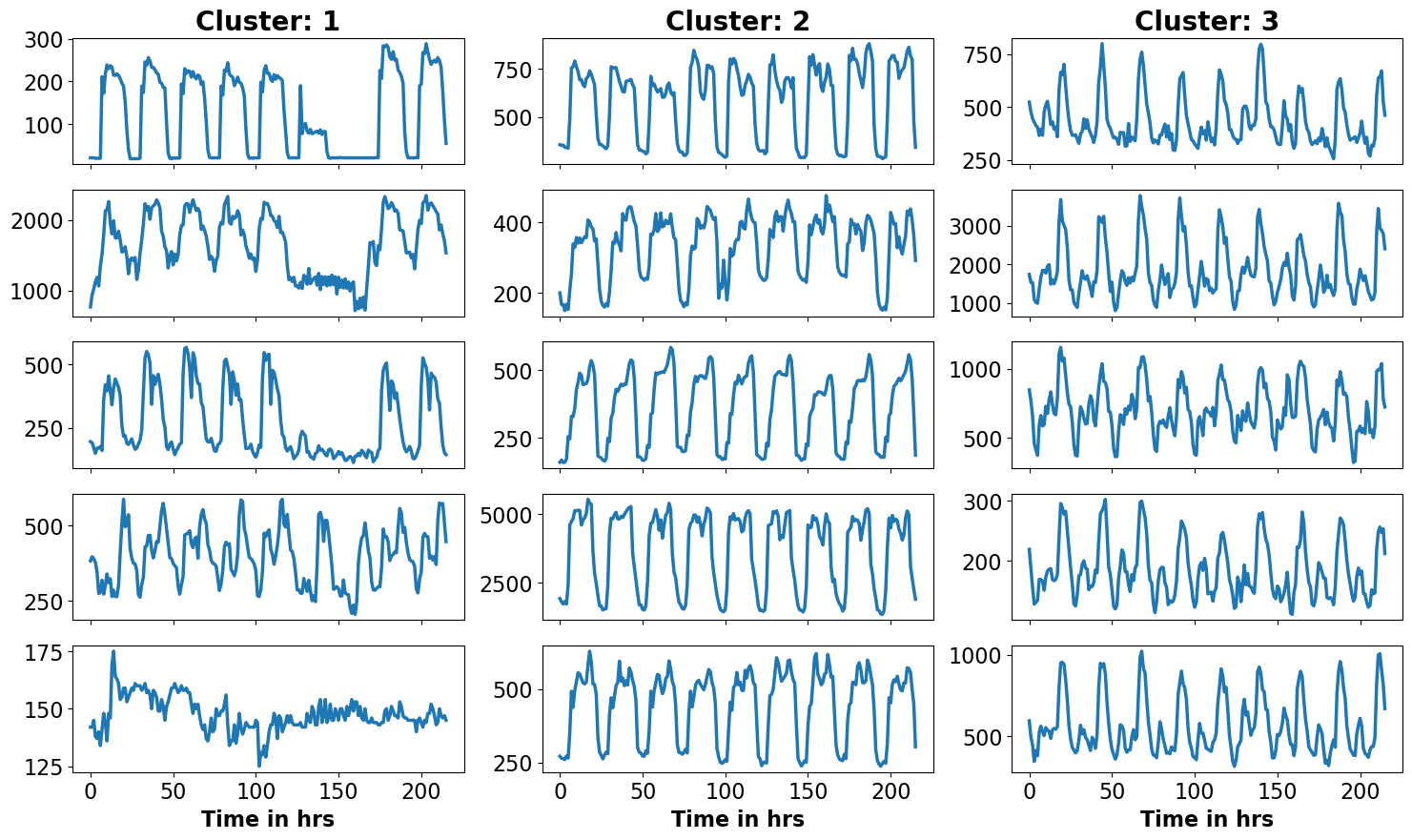}
    \caption{Five example time-series from each cluster of the \textit{Electricity} dataset.}
    \label{fig:ExTrMul}
\end{figure}

We used the \textit{Electricity}\footnote{\url{https://tinyurl.com/fxbcbufp}} dataset \cite{Wu2021Autoformer}\footnote{\url{https://github.com/thuml/Autoformer}} for performance evaluation, which includes hourly electricity consumption of 320 smart meters from July 2016 - July 2019. We selected the first 17,566 entries from July 2016 to July 2018 for training to minimize training time. After normalizing each client's time series to zero mean and unit variance, we used the agglomerative clustering algorithm to group them into three clusters, consisting of 54, 201, and 65 clients, respectively. 
For visualization, five randomly selected examples from each cluster over 9 days of data are shown in Fig. \ref{fig:ExTrMul}. Cluster 1 had the most diverse electricity usage patterns, with load patterns and magnitudes varying significantly across clients in each cluster.

The objective of the proposed framework is thus to learn DL models which can accurately perform predictions for clients from each of the three neighbourhoods. We divide each client's load time series into train-val-test sets for training and evaluation purposes according to a 7:1:2 ratio. {The evaluation metrics selected are the mean absolute error (MAE), mean square error (MSE), and the coefficient of determination ($R^2$) between the true and predicted trajectories. The $R^2$ metric highlights the goodness of fit of a prediction model and is computed as:} 
\begin{equation}
    R^2 = 1 - \dfrac{\sum_i (y_i - \hat{y}_i)^2}{\sum_i (y_i - E[{y}_i])^2},
    \label{eq:r2}
\end{equation}
where $y_i$ is the true time series and $\hat{y}_i$ is the model prediction.
%%%%%%%%%%%%%%%%%%%%%%%%%%%%%%%%%%%
\subsubsection{Implementation Detail} \label{sec:ImpDetail}
All of our experiments were performed on Python 3.8, and model implementations were forked from FEDformers' GitHub implementation\footnote{\url{https://github.com/MAZiqing/FEDformer}}. We keep the default parameters from FEDformer unchanged unless specified otherwise.
Models were trained using ADAM optimizer with adaptive learning rates starting at $10^{-4}$ and a batch size of 32, with MSE as our training loss. Training is performed over 10 epochs, and an early stopping counter of 3 epochs is used to stop the training once the error over the validation set stops to improve. The input sequence length is set at $L=96$ hours, output/prediction horizon is set to $O=96$ hours, model inner dimension $D=512$, and $64$ randomly selected modes are used in FED and FEA blocks. Both FED and FEA are implemented with $h=8$ attention heads.
We choose $K=10$ clients per neighbourhood in 3 neighbourhoods, with one GS assigned to each neighbourhood.
Experiments are repeated 3 times and average results are reported. The split FEDformer contains 2 encoder and 1 decoder layers (see Fig. \ref{fig:Split_Model}).

All DL models are implemented in PyTorch v1.9 \cite{Paszke2019PAi} and training is performed on a single Nvidia Tesla V100-32 GB GPU available through a volta-GPU cluster of the NUS-HPC system\footnote{\url{https://nusit.nus.edu.sg/hpc/}}. To implement differential privacy, we used the open-source \textit{Diffprivlib} library \cite{Holoha2019IBM}. 
The implementation code is publicly available at \cite{SplitCode2022}. 
% \ref{sec:ImpD_App}
% \url{https://github.com/AsifIqbal8739/SplitLoadForecasting}.
% Table generated by Excel2LaTeX from sheet 'Train_Testing_Scores'
\begin{table*}[htbp]
  \centering
  \caption{Test scores for \textit{SplitGlobal} (SL-G), \textit{SplitPersonal} (SL-P), Central, {and FedAvg} models for different GSs' neighbourhoods.}
  % \begin{adjustbox}{width=1\columnwidth}
    \begin{tabular}{l|c|c|c|c||c|c|c|c||c|c|c|c|}
\cmidrule{2-13}          & \multicolumn{4}{c||}{MAE} & \multicolumn{4}{c||}{MSE} & \multicolumn{4}{c|}{$R^2$}\\
\cmidrule{2-13} & {SL-G} & {SL-P} & Central & {FedAvg} & {SL-G} & {SL-P} & Central & {FedAvg} & {SL-G} & {SL-P} & {Central} & {FedAvg}\\
    \midrule
    GS 1  & \textbf{0.451} & 0.453 & 0.513 & {0.535} & 0.387 & \textbf{0.383} & 0.540 & {0.554} & {0.396} & {\textbf{0.405}} & {0.346} & {0.342} \\
    \midrule
    GS 2  & 0.256 & 0.253 & \textbf{0.246} & {0.268} & 0.133 & \textbf{0.130} & 0.133 & {0.142} & {0.863} & {\textbf{0.868}} & {0.855} & {0.838}\\
    \midrule
    GS 3  & 0.372 & 0.361 & \textbf{0.355} & {0.359} & 0.256 & \textbf{0.241} & 0.252 & {0.273} & {0.663} & {\textbf{0.687}} & {0.671} & {0.641}\\
    \midrule
    Mean  & 0.360 & \textbf{0.356} & 0.371 & {0.387} & 0.259 & \textbf{0.251} & 0.308 & {0.323} & {0.641} & {\textbf{0.653}} & {0.624} & {0.607}\\
    \bottomrule
    \end{tabular}%
    % \end{adjustbox}
  \label{tab:OurVsCentral}%
\end{table*}%
%%%%%%%%%%%%%%%%%%%%%%%%%%%%%%%%%%%
\begin{figure*}
    \centering
    \includegraphics[width=1.7\columnwidth]{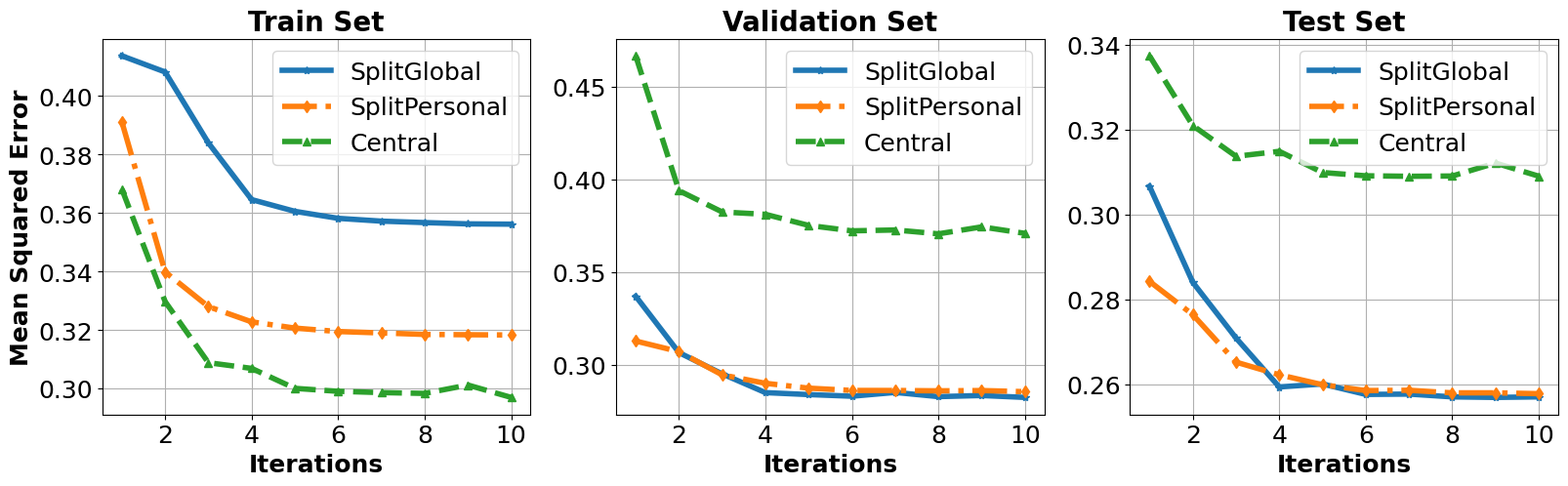}
    \caption{The train, validation, and test set convergence graphs for \textit{SplitGlobal}, \textit{SplitPersonal}, and Central models.}
    \label{fig:ConvG}
\end{figure*}
\subsection{Comparison with Centralized {and FL Models}} \label{sec:CompWCM}
{In our first experiment, we compare the performance of models trained using the proposed SL framework with the FEDformer model trained both centrally and under an FL setting. While FL involves training the entire model on edge clients, in our case, smart meters, this is impractical due to their limited computational capabilities. Nonetheless, we include this comparison for a comprehensive evaluation alongside our approach and the centrally trained model.}

To do so, we train \textit{SplitGlobal} and \textit{SplitPersonal} models for 3 GSs and 10 clients per GS. The clients are kept fixed during all training epochs. For the centralized model, we choose the same 10 clients from each GS (30 clients total), as used by our proposed models to train a univariate FEDformer model in a centralized manner. {The same strategy is used to train a single model in collaboration with 30 clients under FedAvg scheme as well.} The objective is not only to validate our proposed SL training framework but also to assess the efficacy of training multiple models w.r.t. {the single model under central and FL training strategies.} 
The resulting MAE, {MSE, and $R^2$} scores on the test sets of clients belonging to different neighbourhoods are summarized in Table \ref{tab:OurVsCentral}. The MAE, MSE, {and $R^2$} metric scores for GS 2 and 3 for all 3 models are relatively similar, with the Centrally trained model edging the proposed models slightly in terms of MAE, whereas, in terms of MSE {and $R^2$,} the \textit{SplitPersonal} model performing better. 
These results show that all 3 models {could} generalize the load patterns for GS 2 and 3's neighbourhood clients. The load profiles of these clients were not too erratic compared to clients belonging to neighbourhood 1, for whom the centrally trained model performed objectively worse. 

As discussed in Section \ref{sec:Dataset}, the load profiles for clients belonging to cluster 1 (GS 1) show high diversity and are the most challenging out of the three. Observing the superior performance of the proposed models as compared to the central model shows that expecting a single central model to perform well for a diverse range of load profiles is not viable. Moreover, learning multiple models for data from similar distributions should work comparatively well. Thus, the proposed framework essentially offers a balanced approach between learning a single model for all clients and learning a single model for each client.

When comparing scores of \textit{SplitGlobal} and \textit{SplitPersonal}, we see that the latter performs well both in terms of individual GS scores and average ones. This, however, is expected as for \textit{SplitPersonal}, the SP trains personalized Split-2 networks for each GS. Thus it should be able to generalize well for the respective neighbourhoods. To visualize the model convergence, we present the MSE scores for train, validation, and test sets for the 3 models in Fig. \ref{fig:ConvG}.
Here, we see that although the central model achieved the lowest training error, its validation and test set errors are the highest, owing to over-fitting to the training data. We can say this because most, if not all, of the difference between central and proposed methods test set scores, is coming from its prediction scores for the GS 1s' load profiles (see MSE scores given in Table \ref{tab:OurVsCentral}). As otherwise, the test scores for the central model over GS 1 and 2 were very close to the proposed model's scores. The run-times of a single epoch for \textit{SplitGlobal}, \textit{SplitPersonal}, {Central and FL models are 20, 20, 40, and 28} minutes respectively.
%%%%%%%%%%%%%%%%%%%%%%%%%%%%%%%%%%%
\subsection{Across neighbourhood predictions} \label{sec:AcrNP}
\begin{figure}
    \centering
    \includegraphics[width=1\columnwidth]{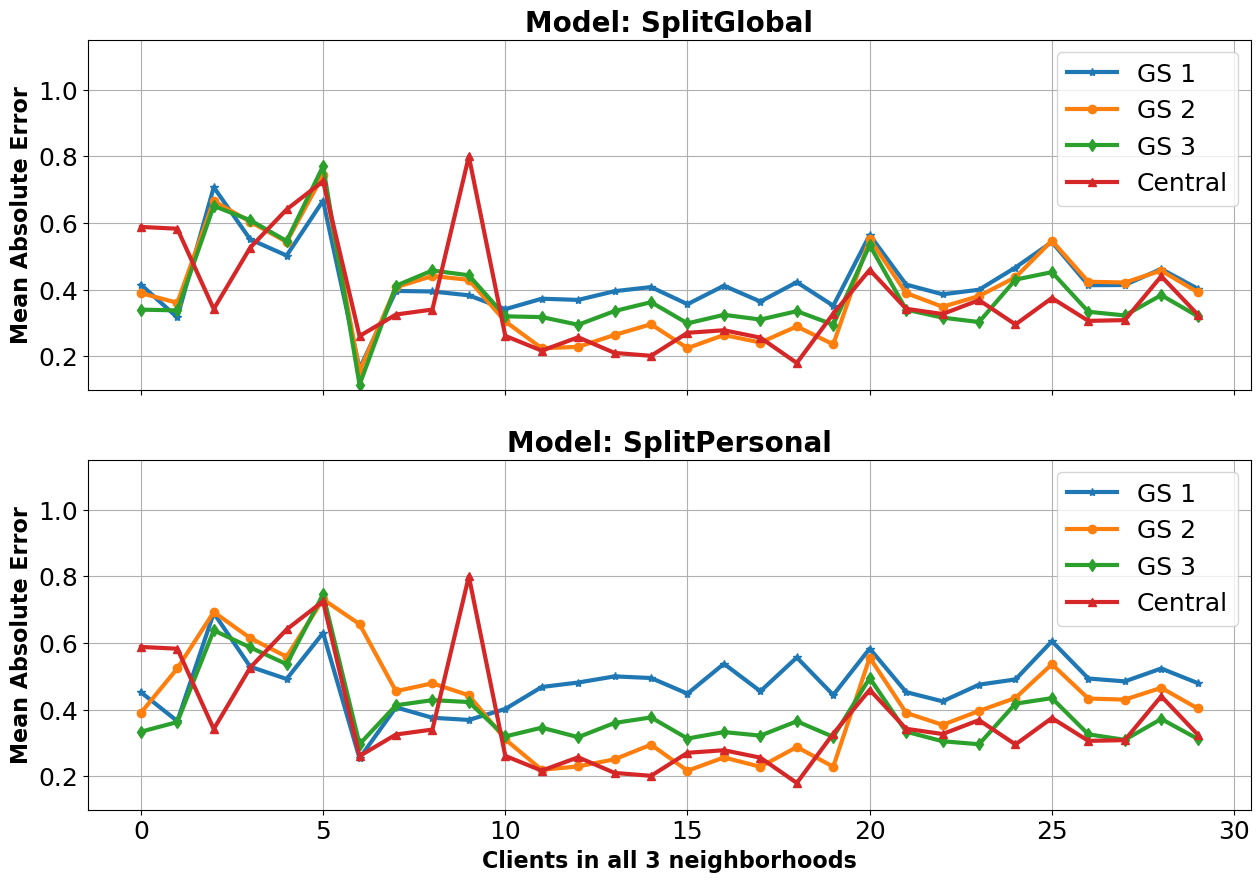}
    \caption{The test prediction errors of every client for \textit{SplitGlobal}, \textit{SplitPersonal}, and Central models.}
    \label{fig:ClErComp}
\end{figure}

In our next experiment, we analyze the trained models' prediction ability when the tested data comes from another GSs' neighbourhood, i.e., from a different distribution. The MSE scores for both models are reported in Table \ref{tab:AcrossNeigh}, where rows GS 1-3 denote the trained split models and columns denote the neighbourhoods. Thus, the table cell for row GS 1 and column 1 represent the MSE score for neighbourhood 1's test data when GS 1's model is used for prediction.
Consider GS 1 models' scores for all neighbourhoods (top row) under both SP training strategies. We see that, apart from testing scores for their own neighbourhood data, the \textit{SplitGlobal} models' prediction errors for neighbourhoods 2 and 3 are well below those reported by \textit{SplitPersonal} model (see Fig. \ref{fig:ClErComp} for individual scores).
% \ref{fig:ClErComp}
This is attributed to the fact that under \textit{SplitGlobal}, the SP learns a single global Split-2 model, which is jointly trained to learn from all neighbourhood clients. Whereas, under \textit{SplitPersonal} training strategy, the SP learns 3 personalized Split-2 models, which have never seen the data coming from different neighbourhoods. Similar trends can be observed for models GS 2 and 3 where cross-neighbourhood scores reported by \textit{SplitGlobal} are better. Similarly, the mean scores for a single neighbourhood overall GS 1-3 models also show that \textit{SplitGlobal} has better generalization capabilities as compared to \textit{SplitPersonal}. 

% Table generated by Excel2LaTeX from sheet 'Cross tests'
\begin{table}[htbp]
  \centering
  \caption{{GS model test scores on clients from all GS neighbourhoods.}}
  \begin{adjustbox}{width=\columnwidth}
    \begin{tabular}{l|c|c|c|c|r|c|c|c|c}
\cmidrule{2-5}\cmidrule{7-10}          & \multicolumn{3}{c|}{SplitGlobal - MSE} &       &       & \multicolumn{3}{c|}{SplitPersonal - MSE} &  \\
\cmidrule{2-5}\cmidrule{7-10}          & 1     & 2     & 3     & Mean  &       & 1     & 2     & 3     & Mean \\
\cmidrule{1-5}\cmidrule{7-10}    GS 1  & 0.390 & 0.249 & 0.353 & \textbf{0.331} &       & 0.385 & 0.379 & 0.434 & 0.399 \\
\cmidrule{1-5}\cmidrule{7-10}    GS 2  & 0.442 & 0.134 & 0.343 & \textbf{0.306} &       & 0.522 & 0.130 & 0.345 & 0.332 \\
\cmidrule{1-5}\cmidrule{7-10}    GS 3  & 0.434 & 0.186 & 0.259 & 0.293 &       & 0.426 & 0.204 & 0.240 & \textbf{0.290} \\
\cmidrule{1-5}\cmidrule{7-10}    Mean  & \textbf{0.422} & \textbf{0.190} & \textbf{0.319} &       &       & 0.445 & 0.238 & 0.340 &  \\
\cmidrule{1-5}\cmidrule{7-10}    
    \end{tabular}%
    \end{adjustbox}
  \label{tab:AcrossNeigh}%
\end{table}%

\begin{figure*}
    \centering
    \begin{tabular}{cc}
         \includegraphics[width=0.8\columnwidth]{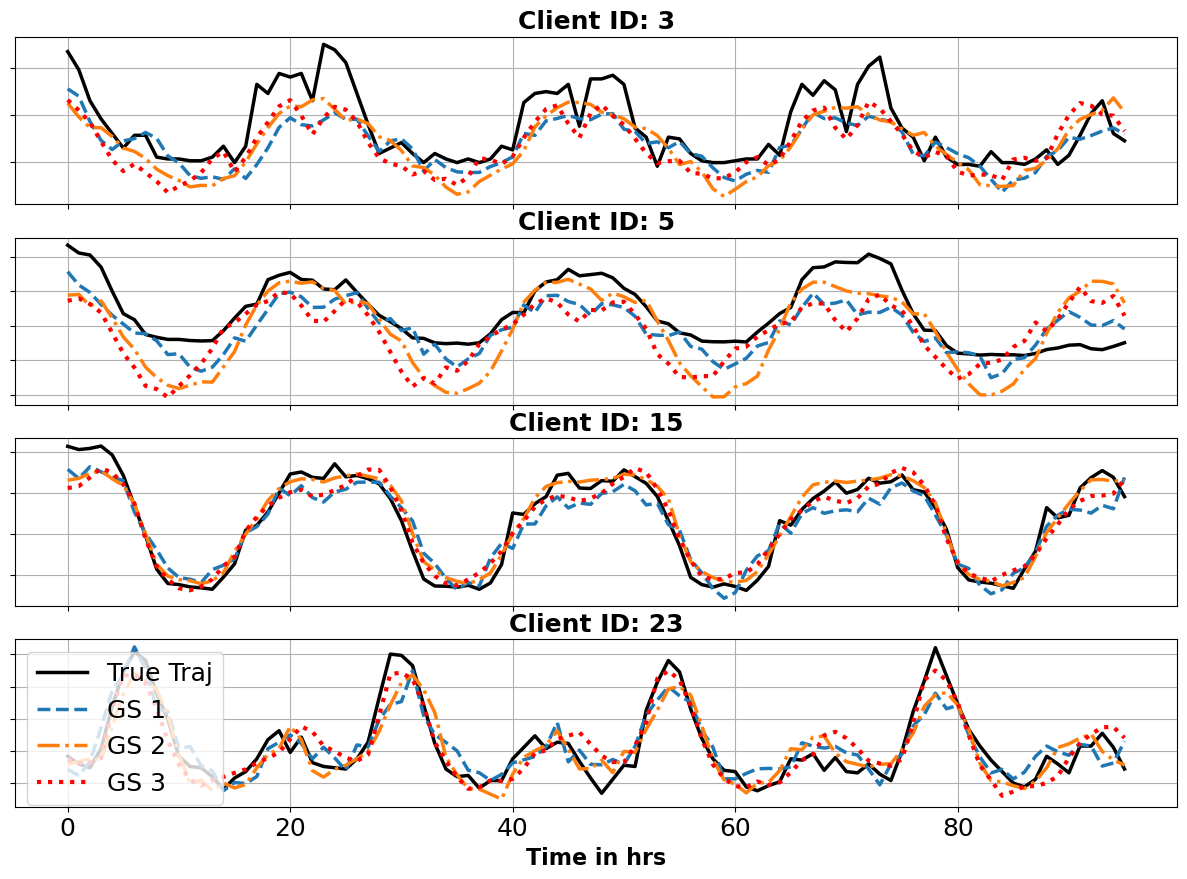} &  
         \includegraphics[width=0.8\columnwidth]{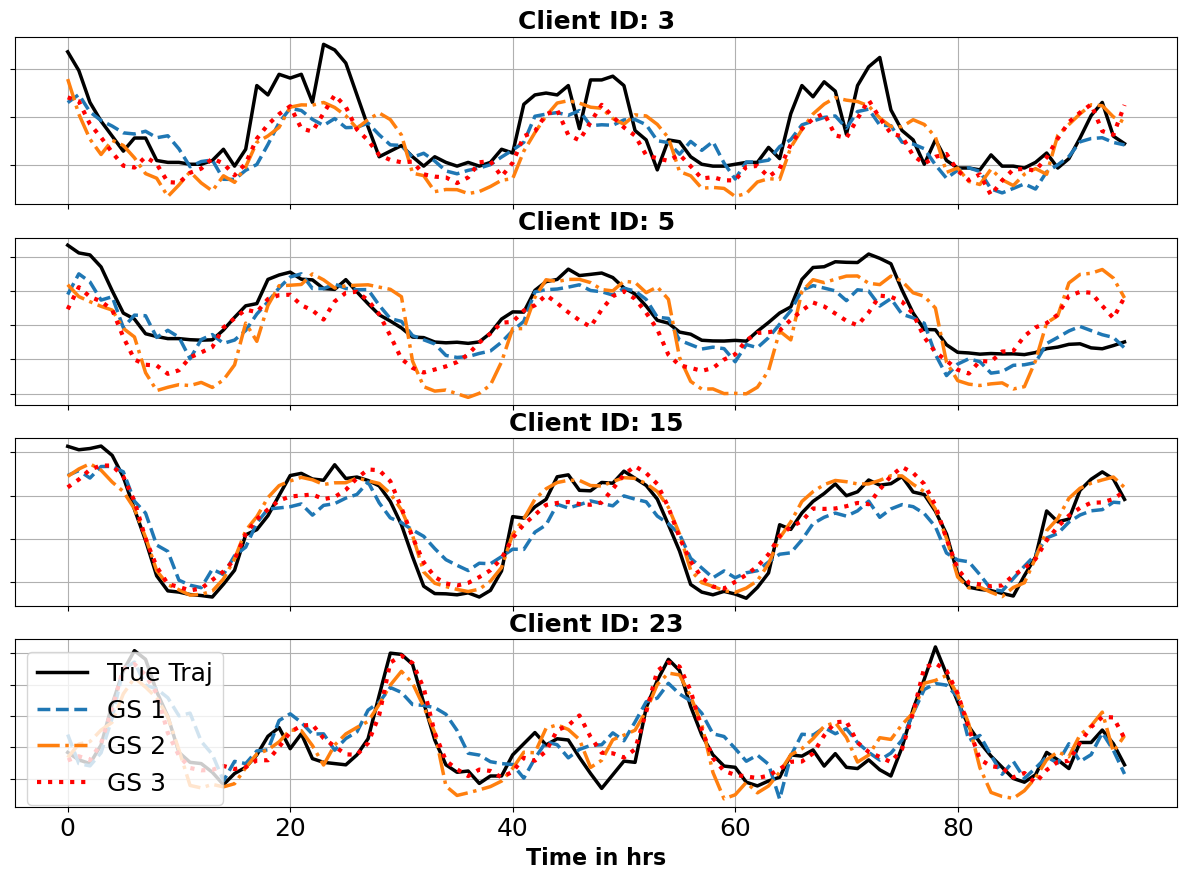} \\
         {SplitGlobal} & {SplitPersonal}
    \end{tabular}
    \caption{A single batch prediction for 4 different clients generated by \textit{SplitGlobal} and \textit{SplitPersonal} over the prediction horizon of 96 hours (4 days).}
    \label{fig:PredVis}
\end{figure*}

To visualize individual test scores for each neighbourhood client under different learned models, we present their MAE scores in Fig. \ref{fig:ClErComp}. 
Here, the first 10 clients are from neighbourhood 1 and so on. For all 3 models, we see that the most MAE variation is found in neighbourhood 1's clients, whereas for 2nd and 3rd neighbourhoods, the variations are small. Moreover, for each neighbourhood, their respective GS models perform best. We further present example batch predictions for 4 clients in Fig. \ref{fig:PredVis}. Looking at predictions from both models, we see that for clients 15 and 23, all three GS models do follow the true trajectory very closely, with \textit{SplitGlobal}'s cross neighbourhood predictions being slightly better. However, for clients 3 and 5, belonging to neighbourhood 1, discrepancies are large, especially for cross-neighbourhood predictions of client 5. Moreover, \textit{SplitPersonal}'s GS 1 model followed the true trajectory much more closely, especially near the prediction endpoints.

% Table generated by Excel2LaTeX from sheet 'Cross tests'
\begin{table}[htbp]
  \centering
  \caption{{Test MSE scores for SL-G and SL-P on seen and unseen data.}}
  \begin{adjustbox}{width=0.8\columnwidth}
    \begin{tabular}{l|c|c|c|c|c|c|}
% \cmidrule{2-7}    \multicolumn{1}{r|}{} & \multicolumn{6}{c|}{Test MSE Scores} \\
\cmidrule{2-7}    \multicolumn{1}{r|}{} & \multicolumn{2}{c|}{{Same Clients}} & \multicolumn{2}{c|}{{Random Clients}} & \multicolumn{2}{c|}{Post Add. Tr.} \\
\cmidrule{2-7}    \multicolumn{1}{r|}{} & SL-G & SL-P & SL-G & SL-P & SL-G & SL-P \\
    \midrule
    GS 1  & 0.390 & 0.385 & 0.650 & 0.644 & 0.528 & 0.511 \\
    \midrule
    GS 2  & 0.134 & 0.130 & 0.166 & 0.161 & 0.154 & 0.147 \\
    \midrule
    GS 3  & 0.259 & 0.240 & 0.218 & 0.295 & 0.214 & 0.264 \\
    \midrule
    Mean  & 0.261 & 0.252 & 0.345 & 0.367 & 0.299 & 0.308 \\
    \bottomrule
    \end{tabular}%
      \end{adjustbox}
  \label{tab:RandomCls}%
\end{table}%
%%%%%%%%%%%%%%%%%%%%%%%%%%%%%%%%%%%
\subsection{Predictions on unseen data} \label{sec:PredUD}
In this experiment, we test the trained models' prediction abilities for clients that are completely new to them. Under usual circumstances, we might not be able to use every client for training; thus, the trained models' generalization capabilities need to be tested. To do so, we start with the models trained on the same 10 clients per neighbourhood and test them on 10 randomly selected clients from each neighbourhood. The results of this test are given in Table \ref{tab:RandomCls}, where we see that clients coming from 2nd and 3rd neighbourhoods receive slightly worse prediction errors, whereas for neighbourhood 1, the error increases by almost 66\%. This performance loss, again, can be attributed to the data diversity found in clients from Neighbourhood 1. To investigate whether further training using these clients brings improvement, we train the model for 5 additional epochs using the selected random clients, whose testing results are summarized in the last two columns of Table \ref{tab:RandomCls}. We see that additional training not only reduced the error for neighbourhood 1 from 66\% to 35\%, but also improved scores for the rest of the neighbourhood clients.
%%%%%%%%%%%%%%%%%%%%%%%%%%%%%%%%%%%
\subsection{Training with random clients} \label{sec:TrRC}
In this experiment, we compare the prediction performance of models when trained using the same clients (10 per GS) compared to randomly selected clients at every training epoch. The models were trained for 10 epochs and tested against randomly chosen clients, and their results are shown in Table {\ref{tab:RandTrain}}. Comparing \textit{SplitPersonal}s' GS 1 performance for both training schemes; we see that scores for the model trained using random clients {are} significantly lower than {those} trained using the same clients. This could be attributed to the fact that in the former case, the model sees several unique clients during the training stage and is thus able to generalize well for unseen data. Apart from this case, the remaining scores are very similar across the two training schemes. From this observation, we can conclude that training using randomly chosen clients every epoch should {enable the models to} perform well when presented with unseen data. However, as discussed in the previous section, performing a few training iterations using the unseen data should be performed to get improved results. 
% Table generated by Excel2LaTeX from sheet 'Cross tests'
\begin{table}[htbp]
  \centering
  \caption{Test MSE scores for random clients with models trained using same vs random clients in every training epoch.}
    \begin{tabular}{l|c|c||c|c|}
\cmidrule{2-5}    \multicolumn{1}{r|}{} & \multicolumn{2}{c||}{Same Clients} & \multicolumn{2}{c|}{Random Clients} \\
\cmidrule{2-5}    \multicolumn{1}{r|}{} & \multicolumn{1}{l|}{SL-G} & \multicolumn{1}{l||}{SL-P} & \multicolumn{1}{l|}{SL-G} & \multicolumn{1}{l|}{SL-P} \\
    \midrule
    GS 1  & 0.650 & 0.644 & 0.638 & 0.551 \\
    \midrule
    GS 2  & 0.166 & 0.161 & 0.164 & 0.157 \\
    \midrule
    GS 3  & 0.218 & 0.295 & 0.224 & 0.294 \\
    \midrule
    Mean  & 0.345 & 0.367 & 0.342 & 0.334 \\
    \bottomrule
    \end{tabular}%
  \label{tab:RandTrain}%
\end{table}%
%%%%%%%%%%%%%%%%%%%%%%%%%%%%%%%%%%%
\subsection{Privacy preservation using differential privacy} \label{sec:PrivAnal}
Under the proposed framework, both network splits are trained outside the clients' premises, and the training is fully controlled by GS and SP entities. Under the honest but curious security assumption, GS and SP shall not deviate from the training process. However, they may try to infer clients' private data from the received Split-1 activations. In order to secure clients' private data, {we propose to use $(\epsilon,\delta)$-DP \cite{Dwork2014Taf, Ji2014}} to safeguard against such inference attacks. However, before we do this, we first analyze information leakage between the input and output of the proposed Split-1 FEDformer model. 
To this end, based on the discussion presented in Section \ref{sec:MINE}, we perform MI estimation using a fully connected neural network (FCNN). A similar strategy has also been used in \cite{Liu2021Aqm} to infer MI between inputs and gradients of an NN under the FL framework.
%%%%%%%%%%%%%%%%%%%%%%%%%%%%%%%%%%%
\subsubsection{Information leakage analysis} \label{sec:InfoLeak}
We use a 3-layer FCNN as our MI neural estimator (MINE), {with layers containing} 100, 50, and 1 neuron, respectively. The first and second layers use the exponential linear unit (ELU) as their activation function. The loss function, given in \eqref{eq:MINE}, is maximized using ADAM optimizer with a learning rate of $10^{-3}$. The batch size is kept at $100$, and the model {was} trained over $10^4$ epochs. Consider an input tensor for Split-1 model $X_{Inp}=concat(X,X_t)$ (see Fig. \ref{fig:Split_Model}) of size $nB\times L \times 5$, where $nB=32$ is the batch size, $L=96$ is the input sequence length, and each time points consists of 1 load value and 4-dimensional date-time encoded vector. The output tensor of Split-1 Encoder block $X_{Out}=X_{enc}^{S_1}$ is of size $nB\times L \times D$, where $D=512$ is the inner model dimension. We aim to find the mutual information leakage between $X_{Inp}$ and $X_{Out}$ for a fully trained Split network.

\begin{figure}
    \centering
    \includegraphics[width=.9\columnwidth]{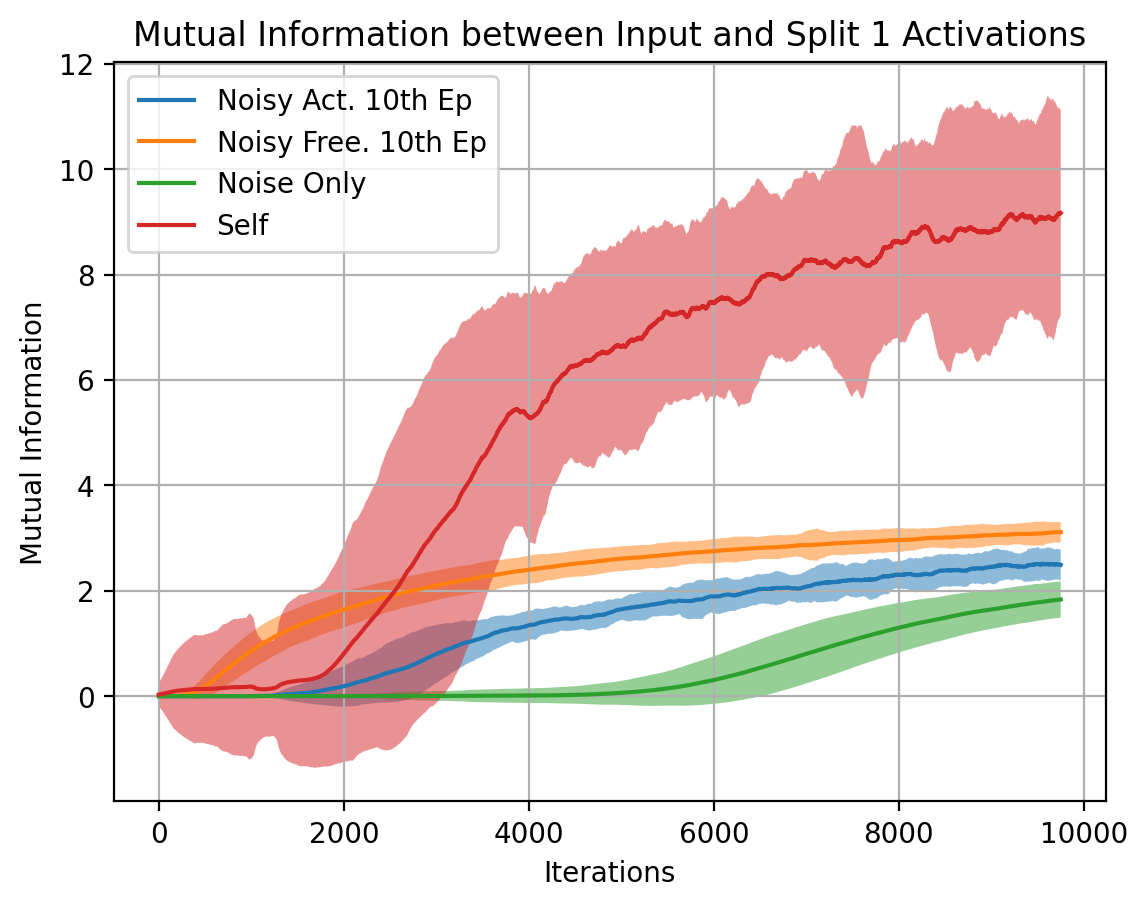}
    % \caption{{The MI between input $X_{Inp}$ and various controlled outputs.}}
    \caption{The mutual information between input $X_{Inp}$, vs itself, Laplace noise only, clean Split-1 activations $X_{Out}$, and $X_{Out}$ plus Laplace noise $\mathcal{L}(b=2)$. The shown spread signifies one standard deviation.}
    \label{fig:MIinfo}
\end{figure}

To establish a baseline, we first train the MINE to find the MI between $X_{Inp}$ and itself using a randomly selected client from neighbourhood 2. In order to reduce the computational time, MI is estimated for every 10th batch. The mean and spread over one std of MI computed for each batch is shown in Fig. \ref{fig:MIinfo} with approximately a final mean MI value $9.12$. The spread seen above and below the mean line signifies the variations of computed MI values over different example batches. With our experimental setting, we can say that the maximum MI between two variables can be at most $9$ on average. Having found the upper limit on MI, we next approximate MI between input and a noise tensor of size equal to $X_{Out}$ to establish a lower limit. The entries of noise tensor were generated from Laplace distribution $\mathcal{L}(b=2)$. The MI score trend for this setting (input - noise only) is also given in Fig. \ref{fig:MIinfo} with a final mean value of approx. $1.83$. Next, we approximate MI between input and clean $X_{Out}$ as well as $\mathcal{L}(b=2)$ noise-contaminated $\tilde{X}_{Out}$ and plot them in the same Fig. \ref{fig:MIinfo} as well. The final MI for clean and noisy outputs were approx. $3.32$ and $2.6$, respectively.

Based on the MI approximations discussed above, we see that the MI between input and clean {and} noisy outputs has a large difference due to the non-linear relationship induced by the Split-1 network. Moreover, the MI of the clean output is much closer to the noise-only case, signifying that the Split-1 network results in minimal information leakage, making the inference attacks much harder to execute \cite{Liu2021Aqm}. Additionally, as the MI can be computed at the client's end, the client can make an informed decision as to whether current batched activations are safe enough for sharing with the GS. 
%%%%%%%%%%%%%%%%%%%%%%%%%%%%%%%%%%%
\subsubsection{{Analysis under $(\epsilon, \delta=0)-$DP}}
{In this section, we analyze the performance impact of $(\epsilon, \delta=0)-$DP (pure DP) on model training and testing. To do so, we select a single client from neighbourhood 2 and train the split model under \textit{SplitPersonal} setting. However, this time, the client uses the Laplace mechanism \cite{Ji2014, Holoha2019IBM} to apply DP to its layer activations (both $X_{enc}^{S_1}$ and $S_{dec}^{S_1}$, see Fig. \ref{fig:Split_Model}) before forwarding to the GS. The level of noise added is controlled by the privacy budget parameter $\epsilon$, where the noise level is inversely proportional to $\epsilon$. Thus, a lower $\epsilon$ results in strong privacy guarantees as compared to higher ones.}
% Thus, $\epsilon$ on a scale of 10 results in weakest privacy to 0 leading to the strongest privacy, however, in this case, the added noise is so large that the data can not be used for training.
\begin{figure}
    \centering
    \includegraphics[width=.9\columnwidth]{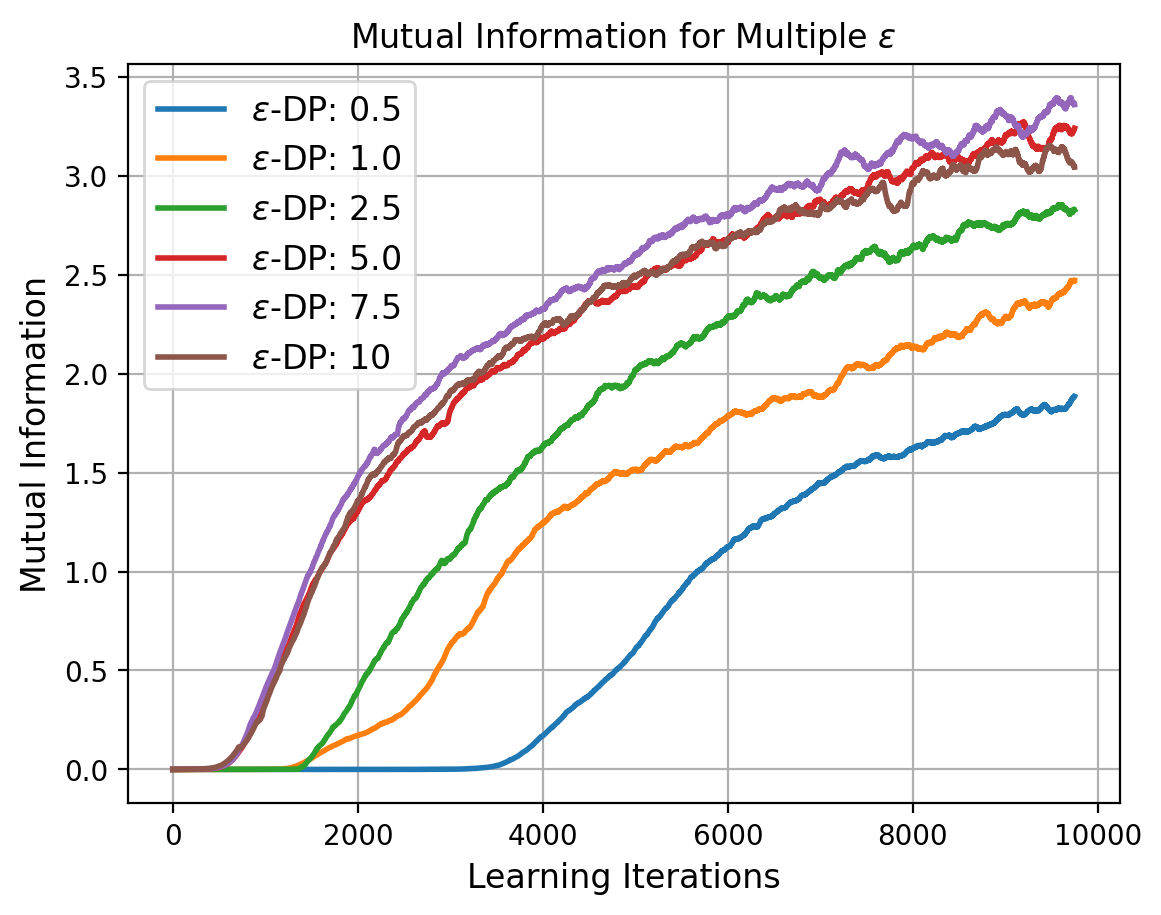}
    \caption{The mutual information learning trend between input and {$(\epsilon, \delta=0)$-DP} protected outputs over a range of $\epsilon$ values.}
    \label{fig:MIEps}
\end{figure}

\begin{figure}
    \centering
    \includegraphics[width=.9\columnwidth]{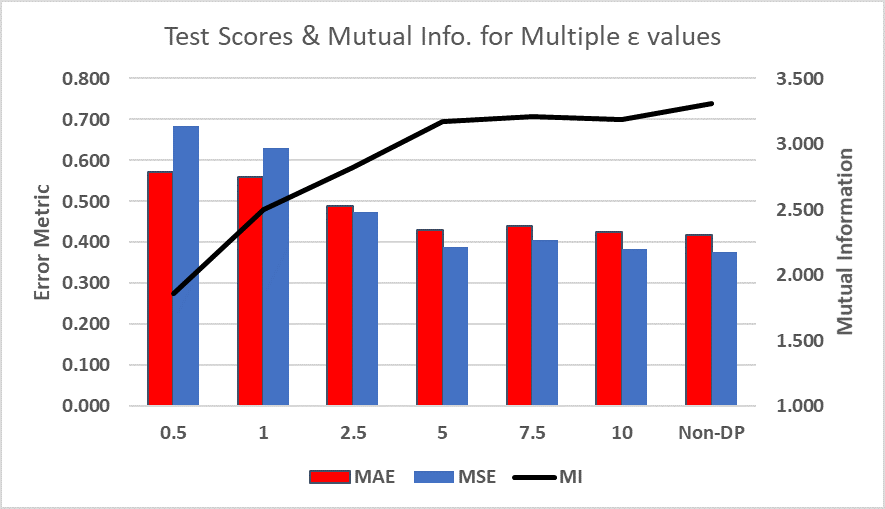}
    \caption{The prediction test scores and mutual information values for models trained with {$(\epsilon, \delta=0)$-DP} over a range of $\epsilon$ values.}
    \label{fig:DPvsMI}
\end{figure}

To analyze the effects of various privacy budgets, we trained the model with $\epsilon\in[0.5, 1.0, 2.5, 5, 7.5, 10]$ and computed the test scores for the models' predictions. At the $10^{th}$ training epoch, we also trained our MINE for every 10th batch to approximate the MI between input and Split-1 layer activations. The MI trends for multiple $\epsilon$ are shown in Fig. \ref{fig:MIEps}, and their respective final MI values, as well as error metrics, are presented in Fig. \ref{fig:DPvsMI}. From Fig. \ref{fig:MIEps} we see that the model trained with $\epsilon=0.5$ has the lowest approximated MI at $1.86$, which is very close to the MI between input and noise only case seen in Fig. \ref{fig:MIinfo}. As a result of large noise additions, its respective MAE and MSE scores are the worst. However, in terms of MAE, they are only 36\% higher than MAE of the non-DP case. Furthermore, increasing $\epsilon$ leads to an increase in MI. However, for $\epsilon\ge 5.0$, the MI stagnates and stays very close to $3.3$, which is the MI of the non-DP case. In this range, the error metrics are within a margin of error to the non-DP case. This shows that the model can handle DP noise for a low $\epsilon$ of 5.0. Even for an $\epsilon=2.5$ (medium privacy), the MAE and MSE are only 16\% and 26\% above the non-DP case, showing that good privacy protection can be achieved with mild performance reduction.
%%%%%%%%%%%%%%%%%%%%%%%%%%%%%%%%%%%
\subsubsection{Analysis under $(\epsilon, \delta)-$DP}
Compared to the stricter pure DP \cite{De2012epDP}, $(\epsilon,\delta)$-DP introduces an extra parameter, $\delta$, into the framework. The reason behind this addition is to provide a level of plausible deniability, allowing for a small probability ($\delta$) that an individual's data might be exposed or identified by an attacker. While $\epsilon$ governs the average privacy loss incurred, $\delta$ plays a role in controlling the worst-case privacy loss scenario. Another advantage of $(\epsilon, \delta)-$differential privacy is its advanced composition theorems, enabling significantly more training iterations than pure DP under the same $\epsilon$. This is the reason why most related works in differentially private machine learning have shifted away from pure differential privacy.
\begin{figure}
        \centering
        \includegraphics[width=.9\columnwidth]{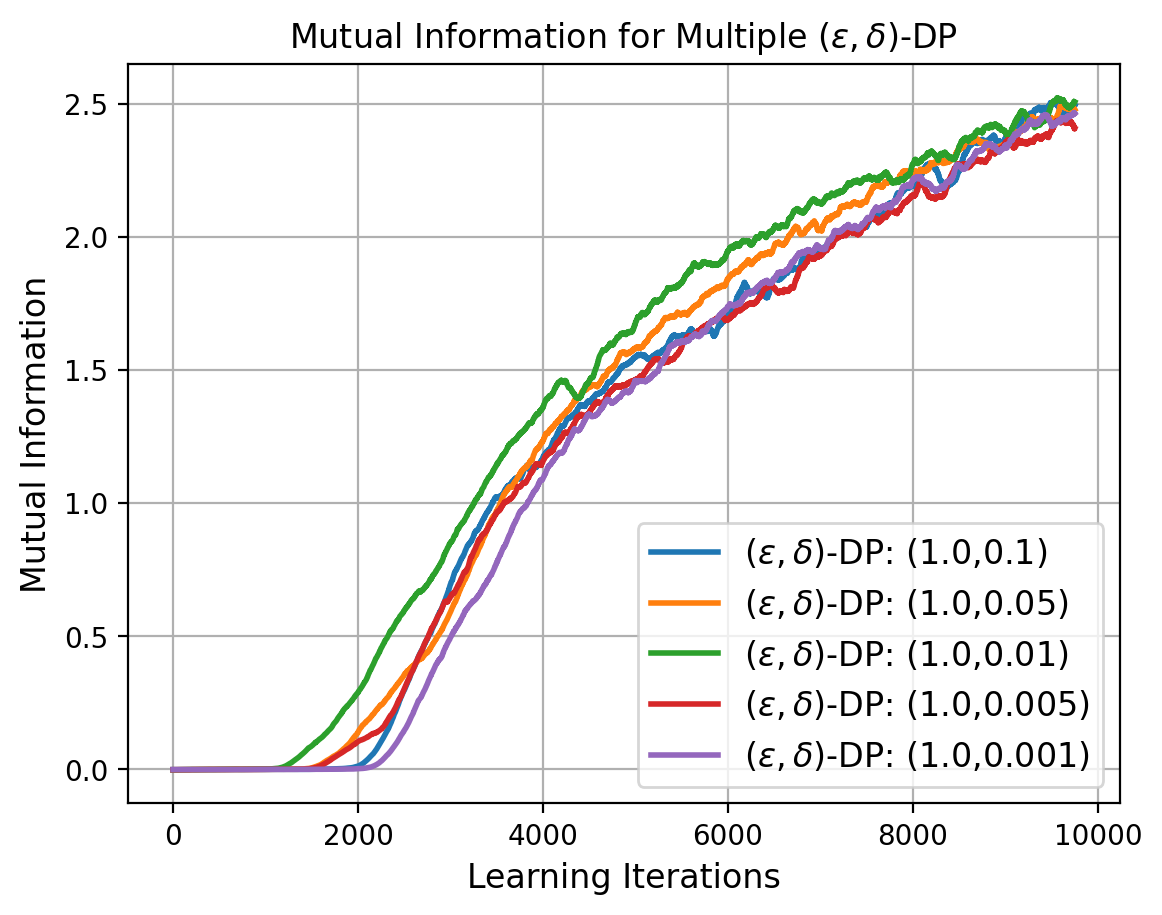}
        \caption{The mutual information learning trend between input and $(\epsilon,\delta)$-DP protected outputs over fixed $\epsilon=1.0$ and a range of $\delta$ values.}
        \label{fig:MIEps_2}
\end{figure}

\begin{figure}
    \centering
    \includegraphics[width=0.9\columnwidth]{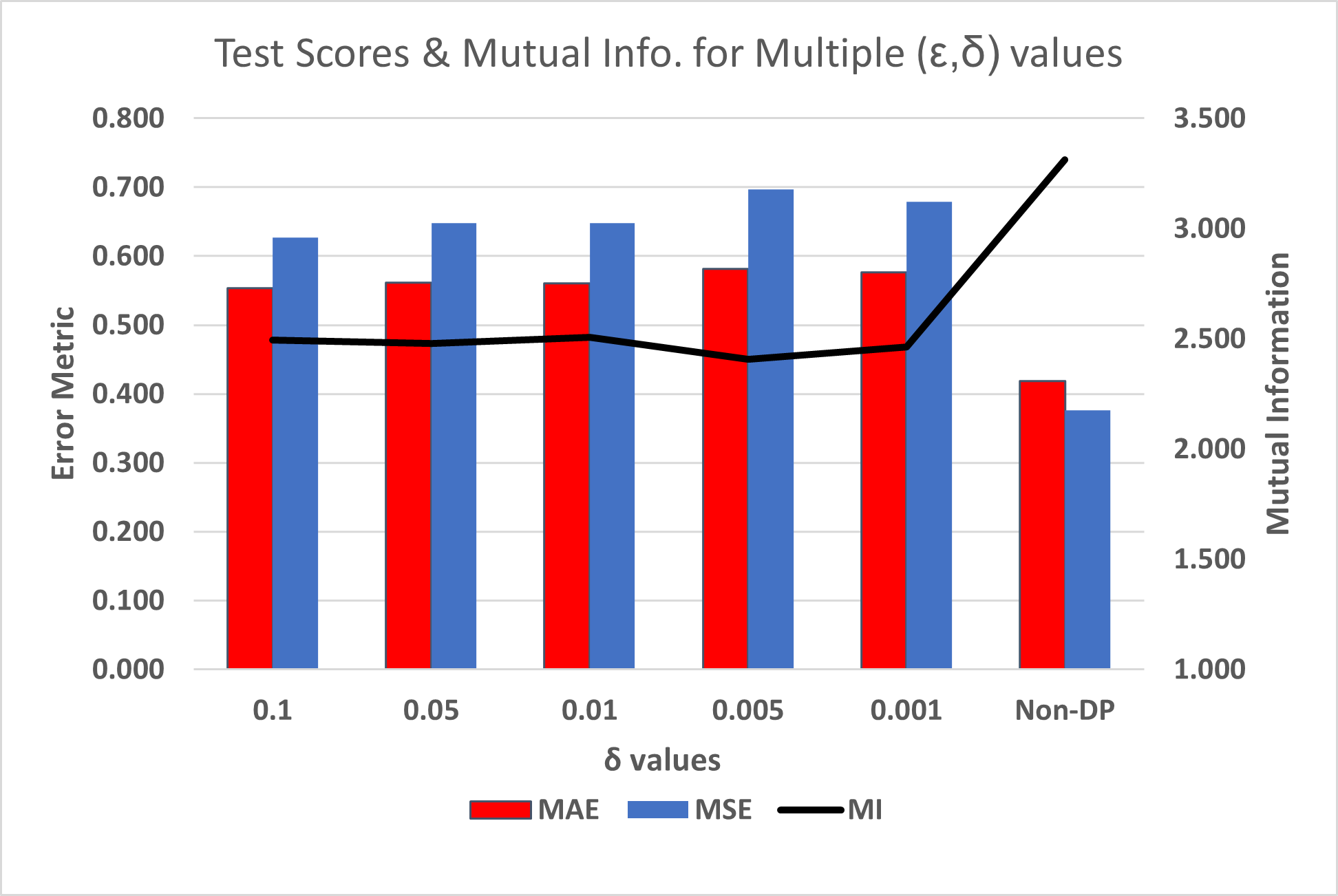}
    \caption{The prediction test scores and mutual information values for models trained with $(\epsilon,\delta)$-DP over fixed $\epsilon=1.0$ and a range of $\delta$ values.}
    \label{fig:DPvsMI_2}
\end{figure}

To investigate the impact of $(\epsilon,\delta)$-DP on MI leakage and model performance, we repeat the previous experiment but with the $(\epsilon,\delta)$-DP via Gaussian mechanism \cite{Holoha2019IBM}. In this experiment, we maintained $\epsilon$ at a fixed value of $1.0$ and systematically varied $\delta$ within the range $[0.1, 0.05, 0.01, 0.005, 0.001]$, representing a transition from higher worst-case privacy loss to lower levels. The resultant MI leakage, estimated by our estimator (MINE), between clients' input data and Split-1 layer activations is presented in Figure \ref{fig:MIEps_2}. Interestingly, the variations in $\delta$ had negligible effects on average privacy loss, as $\delta$ primarily governs the worst-case scenario. Furthermore, Figure \ref{fig:DPvsMI_2} displays the model's prediction accuracy over the entire range of $\delta$ values. These metrics indicate that neither MAE nor MSE exhibit substantial variations within the considered range of $\delta$ values. {The performance closely} mirrors the outcomes observed when $\epsilon=1.0$, as depicted in Figure \ref{fig:DPvsMI}.

\begin{figure}
    \centering
    \includegraphics[width=0.9\columnwidth]{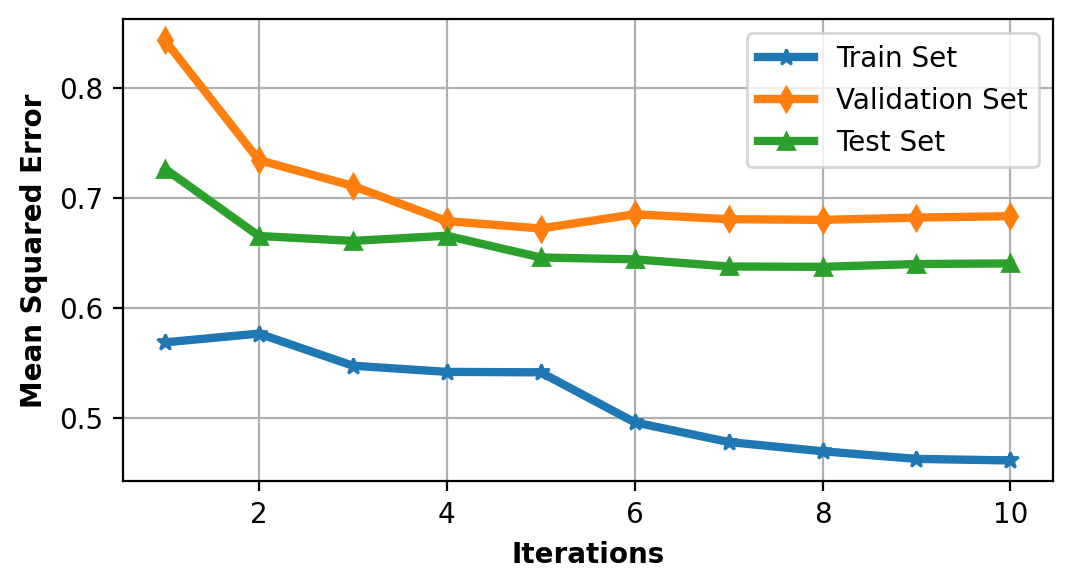}
    \caption{{The convergence graphs for train, validation, and test sets under $(\epsilon=1.0, \delta=0.005)$-DP setting.}}
    \label{fig:ConvGraphs_DP}
\end{figure}

{To illustrate the impact of DP on convergence behavior, Fig. \ref{fig:ConvGraphs_DP} presents the individual loss trends for the training, validation, and test sets over 10 epochs. The plot shows that validation and test losses decrease until about the 5th epoch and then stabilize, while the training loss continues to decline until around the 9th epoch. As expected, the overall loss floor in the DP setting is noticeably higher than in the non-DP case (Fig. \ref{fig:ConvG}) due to the added noise. Additionally, training time increased significantly, from roughly $200$ minutes without DP to approximately $454$ minutes with DP, representing a $2.54 \times$ increase, primarily due to the computational overhead introduced by noise computations performed through the \textit{Diffprivlib} library \cite{Holoha2019IBM}.}

%%%%%%%%%%%%%%%%%%%%%%%%%%%%%%%%%%%
\subsection{{Feature Space Hijacking Attack}} \label{sec:FSHA}
Although our framework is built under the `honest-but-curious' assumption, i.e., the utility provider, which is in control of both GS and SP entities,  does not intentionally attempt to reconstruct client data or poison the training process, we deliberately relax this assumption to evaluate the worst-case scenario. In this setup, the utility provider (or a compromised entity within it) can deviate from the standard training procedure and attempt to reconstruct private client input data from the shared Split-1 activations using the powerful Feature-Space Hijacking Attack (FSHA) \cite{Pasqui2021Utt}.

\begin{figure}
    \centering
    \includegraphics[width=0.9\columnwidth]{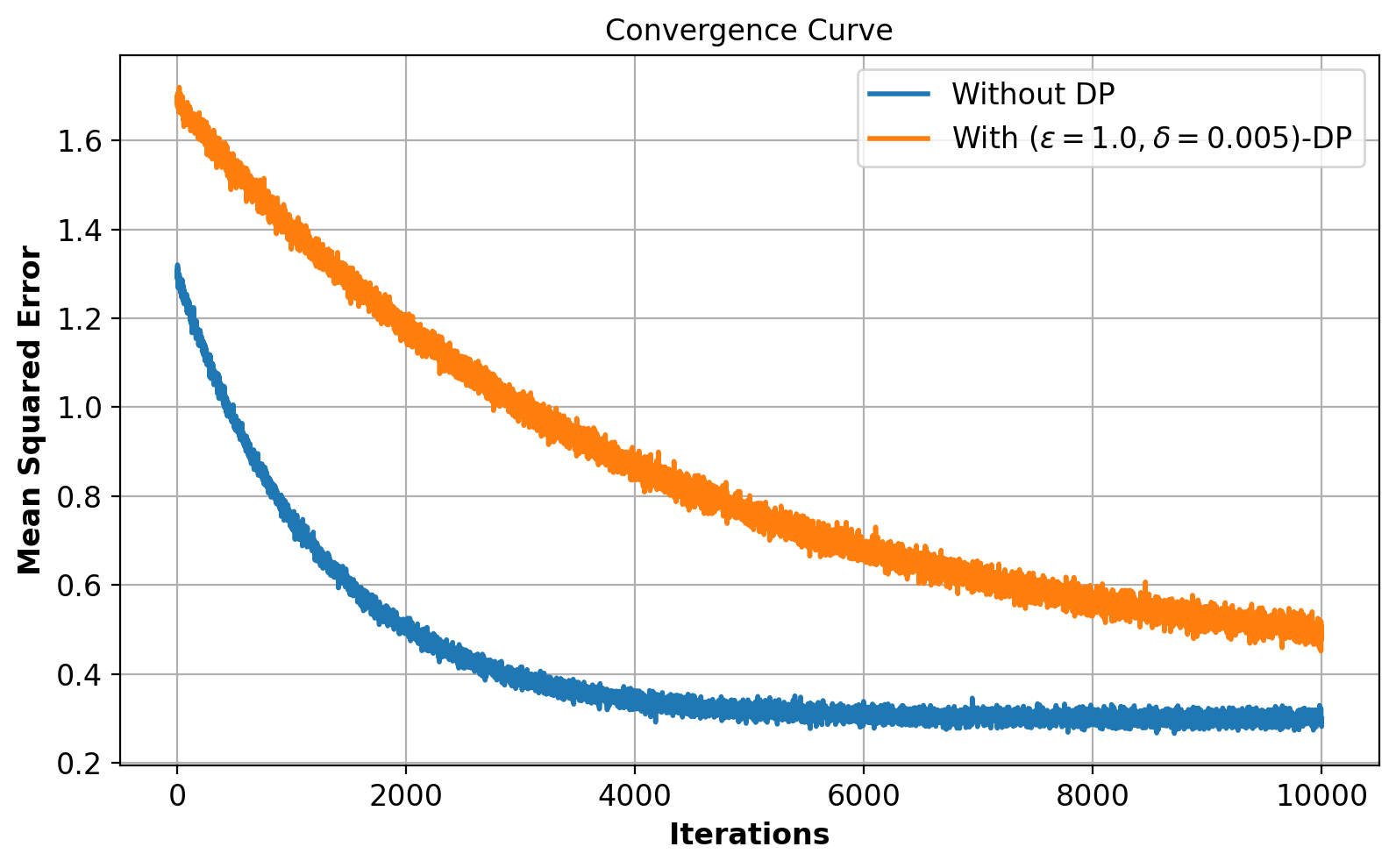}
    \caption{{The reconstruction MSE of the private client data by the FSHA on \textit{SplitGlobal} model with and without DP.}}
    \label{fig:ConvGraphs_FSHA}
\end{figure}

To compare the effects of DP, we apply the FSHA attack on \textit{SplitGlobal} model under two scenarios: with and without DP applied to the client side activations. This experiment uses the same settings as used in Section IV-E3. Under this scenario, the smoothed reconstruction MSEs achieved by the FSHA are illustrated in Fig. \ref{fig:ConvGraphs_FSHA}.

\begin{itemize}
    \item \textbf{Without DP}, the attack model converges faster and achieves lower reconstruction MSE scores, consistent with observations from previous works \cite{Pasqui2021Utt, Gawron2022}. However, even in this case, our proposed framework maintains a reconstruction MSE floor around $0.305$, indicating that the attacker still requires considerable effort to achieve meaningful reconstruction, partly due to the preserved complexity in the input-output relationship at Split-1.
    \item \textbf{With DP}, the attack becomes significantly more difficult. The attacker needs a much longer training period, and the resulting reconstruction accuracy is further degraded. This aligns with the intended purpose of DP: to degrade information leakage and protect against reconstruction-based attacks.
\end{itemize}
The above results are consistent with our analysis under Mutual Information (MI) estimation (section \ref{sec:InfoLeak}) as well. These results demonstrate that our proposed framework, especially when combined with differential privacy, exhibits improved robustness against powerful feature-space attacks. 

%%%%%%%%%%%%%%%%%%%%%%%%%%%%%%%%%%%
\subsection{{Computational and Communication Overhead}} \label{sec:Overhead}
In this section, we summarize our findings regarding the various overheads related to our training framework. For details, the reader is referred to the Appendix \ref{app:A}.
\paragraph{Communication Overhead} In our framework, the majority of communication overhead is handled between the GS and SP components, with minimal burden placed on the client. For each model update round, the client transmits activations from Split-1, amounting to approximately $6.3$ MB, to the GS and receives a negligible $3$ KB prediction output in return. During backpropagation, the client sends back loss gradients of the same size ($3$ KB), while the heavier $63$ MB gradient tensors are exchanged between the GS and SP. As a result, the client is responsible for only about $4.55$\% of the total communication volume, highlighting the communication efficiency of our split learning design. Appendix \ref{app:A1} provides further numerical details.
\paragraph{Computational Overhead} The computational workload in our split learning framework is unevenly distributed between the client (smart meter) and the utility provider (GS-SP). We approximated the number of multiplications for each model component, drawing from the FEDformer architecture. Our analysis shows that the smart meter (Split-1) is responsible for only about $12.66$\% of the total forward pass operations, while the remaining 87.34\% are executed by the SP (Split-2). This is due to the client's role being limited to the initial encoding and decoding blocks, while the heavy attention and feed-forward layers are managed by the utility provider. Additionally, the backpropagation and model updates are entirely performed by the GS-SP entities. Since the backpropagation step generally involves around $2.5\times$ (average of $2-3\times$) the number of forward-pass operations, the client’s total share of computations across a full model update round amounts to approximately $5$\%. This lightweight requirement affirms that smart meters can comfortably participate in training without a significant computational burden. A detailed breakdown is provided in Appendix \ref{app:A2}.
\paragraph{Latency Overhead} Due to the architectural split, most of the computational operations are handled by the GS and SP, which collectively perform over $95$\% of the total workload. Consequently, the latency of each training iteration is primarily determined by the utility provider’s infrastructure. Assuming higher-performance hardware at the GS-SP end, the total round-trip latency remains low. While our implementation was done on a single machine, Appendix \ref{app:A3} provides a theoretical latency analysis to offer real-world latency estimates under reasonable hardware assumptions.
\paragraph{Energy Overhead} The client device in our split learning setup is responsible for approximately $5$\% of the total computational workload during each model update. This offloading significantly reduces energy consumption at the edge, making the approach attractive for power-constrained devices like smart meters. As shown in Table \ref{tab:OurVsCentral}, our SplitPersonal model achieved an $18.5$\% improvement in prediction performance over the centralized baseline while shifting most of the energy burden to the utility provider's infrastructure. A deeper analysis of energy distribution and trade-offs is provided in Appendix \ref{app:A4}.
%%%%%%%%%%%%%%%%%%%%%%%%%%%%%%%%%%%
\subsection{Discussion} \label{sec:Disc}
% With the rise in data privacy and security concerns, the need to develop ML models and frameworks which are capable of training without requiring clients to share their private data. Federated and Split learning frameworks take an important step in making this happen, however, they need to be further analyzed in various contexts. In this study, we presented a split learning approach for the problem of electricity load forecasting. 
In Section \ref{sec:ExpEval}, we used two SL strategies, \textit{SplitPersonal} where we train neighbourhood-level personalized split networks, and \textit{SplitGlobal} where a single global Split-2 is trained at SP for all personalized Split-1 models at GSs. Networks trained under both strategies achieved better or comparable performance compared to a centrally trained network, as seen in Table \ref{tab:OurVsCentral}. When predicting across neighbourhood clients, \textit{SplitGlobal} model was able to get lower errors as compared to \textit{SplitPersonal}, as shown in Table \ref{tab:AcrossNeigh}, which is expected. 

In Section \ref{sec:PredUD}, we tested the trained models on data from clients not used during the training stage. The results in Table \ref{tab:RandomCls} show that the trained models performed well in this scenario; however, additional training using the new client's data improved performance. This is essential as new clients are constantly added to the system, and we might have very little data on them. Additionally, in Section \ref{sec:TrRC}, we compare the performance of models trained on the same clients vs training on random clients every epoch against unseen data and found that the models trained on random clients performed better. As with the availability of large amounts of smart meter data, training using all of it is often not feasible. Instead, training using a random subset of clients every epoch can lead to a model with good generalization capabilities. Furthermore, this model can be refined using unseen clients' data for added performance.

In Section \ref{sec:PrivAnal}, we analyzed the extent of privacy leakage arising from the sharing of clients' activations using MINE. In Fig. \ref{fig:MIinfo}, we showed that even without the added noise, the Split-1 activations have significantly low MI w.r.t. the inputs due to their non-linear and complex relationship induced by {the} Split-1 model. Moreover, based on the estimated MI, a client can decide whether to forward the current batch activations to GS or not to mitigate privacy leakage. We further analyze the effects of introducing differential privacy as an additional layer of security on the model's performance. In Fig. \ref{fig:DPvsMI}, we see that with a moderate privacy budget of $\epsilon=5.0$, the models performed similarly to the non-DP case and saw performance degradation only when $\epsilon\le2.5$, leading to strong privacy. With a trained model, the electricity service provider can perform individual-level predictions (requiring respective clients' involvement) and neighbourhood-level predictions using the cumulative load trend from the grid station servicing the neighbourhood. 
% The LSTM-based FL models \cite{Fekri2022Dlf, Taik2020Elf} proposed so far were able to make 1h to 24h ahead predictions; our transformer-based model can perform accurate predictions from 24h - 720h \cite{Zhou2022FED}, although our experiments were limited to 96h ahead predictions only.

In Section \ref{sec:FSHA} we evaluated the \textit{SplitGlobal} framework under FSHA \cite{Pasqui2021Utt} in both non-DP and DP-enabled settings. In the non-DP case, the attacker's reconstruction achieved an MSE of 0.241, while enabling DP increased the reconstruction MSE to 0.328, indicating a clear improvement in privacy protection through DP regularization. Finally, in Section \ref{sec:Overhead}, our overhead analysis shows that the client-side in the Split learning framework contributes only approx. $5$\% of the total computational load and $4.55$\% of the communication load per training round. Additionally, under reasonable hardware assumptions, the client accounts for approximately $21$\% of the end-to-end latency and similar proportions in energy consumption, highlighting the lightweight nature of the client-side operations.

Our proposed split learning framework is well-suited for integration into modern power systems, particularly within infrastructures like Advanced Metering Infrastructure (AMI) and Energy Management Systems (EMS). The Split-1 module can be embedded in smart meters, while the GS and SP components can be deployed at the utility’s backend, leveraging existing EMS or Distribution Management System (DMS) facilities. This decentralized architecture supports scalable and privacy-preserving model training across diverse consumer endpoints. However, practical deployment may face challenges such as communication latency, device heterogeneity, synchronization issues, and secure data exchange. Addressing these challenges will be essential to fully realize the framework’s potential in real-world settings, and we identify these areas as important directions for future work.
%%%%%%%%%%%%%%%%%%%%%%%%%%%%%%%%%%%
%%%%%%%%%%%%%%%%%%%%%%%%%%%%%%%%%%%
%%%%%%%%%%%%%%%%%%%%%%%%%%%%%%%%%%%
\section{Conclusion} \label{sec:Conc}
In this article, we propose a split learning framework to train a DL time series prediction model using the client smart meter data without compromising individual clients' privacy.
Our proposed SL frameworks use the smart meters to perform a forward pass through the Split-1 network only, while the rest of the training is relegated to GS and SP entities. This ensures that the smart meter's main functionalities remain unhindered.  
Once trained, the energy provider retains the entire model, which can be used to perform load predictions for a single smart meter or the entire neighbourhood. 
The experimental results have shown that the performance of the trained models is better or on par with that of a centrally trained model. 
To analyze the extent of information leakage through the Split-1 network, we used mutual information neural estimation to approximate the MI between the input and output of the Split-1 network. The analysis showed that the MI leakage through the Split-1 network is limited. 
Furthermore, as an added layer of security, we analyzed the addition of $\epsilon$-DP {and $(\epsilon,\delta)$-DP} to our framework under multiple privacy budgets. We found that the models performed similarly to the non-DP case under a medium privacy budget while observing low-performance degradation under relatively low privacy budgets. 

%%%%%%%%%%%%%%%%%%%%%%%%%%%%%%%%%%%
%\paragraph{Acknowledgement} We would like to acknowledge that computational work involved in this research work was fully supported by NUS IT’s Research Computing group.
%================================================================================================
%================================================================================================
% 
%------------------------------------------------
\bibliographystyle{IEEEtran}
\bibliography{SplitLoadForecasting}
\appendices
\section{Comprehensive Overhead Analysis of the Proposed Split Learning Framework}\label{app:A}
To provide a detailed overhead analysis of the proposed framework, we refer to the Dual Split FEDformer model illustrated in Fig. \ref{fig:Split_Model}. The relevant model parameters and their values are listed in Table \ref{tab:Mod_Params}.

In the discussion that follows, we assume a batch size of $1$, which aligns with the per-client data processing during a single training epoch. This choice also reflects the trade-off between training speed and the computational and communication capabilities available at the client (smart meter) side. The analysis below is focused on a single round of model training.

% Table generated by Excel2LaTeX from sheet 'Overhead'
\begin{table}[htbp]
  \centering
  \caption{Model Parameters}
    \begin{tabular}{llr} \toprule
    Symbol & Description & Value \\\midrule
    $C$   & Clients   & 10 \\
    $B$   & Batch Size    & 1 \\
    $N$   & FFT Size    & 128 \\
    $D$     & Model dimension  & 512 \\
    $D_{ff}$  & FFN Dimension & 2048 \\
    $M$     & Modes of FFT kept & 64 \\
    $L_e$    & Encoder Length & 96 \\
    $L_d$    & Decoder Length (1.5 x $L_e$) & 144 \\ 
    $O$ & Prediction Horizon & 96 \\ \bottomrule
    \end{tabular}%
  \label{tab:Mod_Params}%
\end{table}%

\subsection{Communication Overhead:} \label{app:A1}
We break down the communication overhead into forward and backward pass components:
\begin{itemize}
    \item \textbf{Forward Pass:} Each smart meter handles the forward pass through the Split-1 block, computes the final loss, and initiates gradient backpropagation. The set of Split-1 activations sent from each client to the GS, denoted $\textbf{A}_{k,b}^{GS}$ in Algorithm 1, includes the output tensors from the:
    \begin{itemize}
        \item Split-1 Encoder: $X_{enc}^{S_1} \in \mathbb{R}^{B\times L_e \times D}$
        \item Split-1 Decoder: $S_{dec}^{S_1} \in \mathbb{R}^{B\times L_d \times D}$
        \item Trend component from Split-1 Decoder: $trend_{enc}^{S_1} \in \mathbb{R}^{B\times L_d \times D}$.
    \end{itemize}
    Using the model parameters in Table \ref{tab:Mod_Params} and assuming \texttt{Float32} precision, the payload size per client for these activations is approximately $6.3$ MB. The GS aggregates the activations from all $C=10$ clients and transmits them to the SP, resulting in a total payload of approximately $63$ MB.
    
    For loss computation, the SP returns a small output tensor of size $B \times O$ (roughly $3$ KB per client) back to the respective smart meters via the GS, this is negligible in terms of communication overhead.
    
    \item \textbf{Backward Pass:} The client initiates backpropagation by sending the gradient of the loss (approx. $3$ KB) to the SP via the GS. The SP continues backpropagation, computes gradients for Split-1, and sends them (approx. $63$ MB) to the GS. The GS completes the backpropagation and updates the Split-1 model, while the SP updates the Split-2 model. 

    In total, the communication volume during a single model update iteration is: $6.3$ MB (client $\rightarrow$ GS) + $63$ MB (GS $\rightarrow$ SP) + $63$ MB (SP $\rightarrow$ GS) = $132.3$ MB

    Notably, only \underline{$\textbf{4.55}$\%} of the total communication is from the client to the GS; the majority occurs between GS and SP. Please note that these calculations exclude transmission protocol overhead.
\end{itemize}

\subsection{Computational Overhead:} \label{app:A2}
Estimating the total computational cost for a single round of model training is inherently complex due to backend optimizations in frameworks such as PyTorch. However, we can approximate the number of operations by counting the number of multiplications (ignoring additions) performed within key modules of the FEDformer model. These estimates are based on the methodology provided in Section 4.2 of \cite{Zhou2022FED}.

The most computationally intensive modules are the \textit{Frequency Enhanced Block (FEB)}, \textit{Frequency Enhanced Attention (FEA)}, and \textit{Feed Forward Network (FFN)}. The \textit{Series Decomposition Block}, which uses lightweight convolutional operations, is negligible in comparison and is omitted from the analysis. All notations follow those in Table \ref{tab:Mod_Params}.

Module-wise multiplication count:
\begin{itemize}
    \item \textbf{FEB:} 
    \begin{itemize}
        \item Encoder: $\approx\left[(L_e + M)D^2 + 2NL_eD\right]$,
        \item Decoder: $\approx\left[(L_d + M)D^2 + 2NL_dD\right]$.
    \end{itemize}
    \item \textbf{FEA:} $\approx \left[(2L_e+L_d)D^2 + 2(L_e+L_d)ND + 2M^2D \right]$.
    \item \textbf{FFN:} 
    \begin{itemize}
        \item Encoder: $\approx 2L_eDD_{ff}$,
        \item Decoder: $\approx 2L_dDD_{ff}$
    \end{itemize}
\end{itemize}
Using these expressions, we now compute the computational load distribution between the client and the GS-SP infrastructure for a single forward pass.
\begin{itemize}
    \item \textbf{The Client Split-1:} As shown in Fig. \ref{fig:Split_Model}, Split-1 comprises two FEB blocks (one in Encoder, one in Decoder), and a projection via matrix $W_1 \in \mathbb{R}^{D \times Z}$. For our univariate case ($Z = 1$), the projection involves $L_dD$ multiplications. Substituting $L_d = 1.5L_e$ and using $L_e = L$, the total number of multiplications becomes:
    \begin{equation}
        \textrm{Ops}_\textrm{Client} = (2.5L+2M)D^2 + (5N+1.5)LD 
    \end{equation}
    which evaluates to approximately $1.28 \times 10^8$ multiplications. 
    \item \textbf{The SP Split-2:} The SP side consists of:
    \begin{itemize}
        \item Encoder Layers: 2 FFN modules and 1 FEB
        \item Decoder Layer: 1 FEA block, 1 FFN block, and 3 projection operations
    \end{itemize}
    The total multiplication count is:
    \begin{equation}
        \textrm{Ops}_\textrm{SP} = (4D_{ff}+7N+4.5)LD + (4.5L+M)D^2 + 2M^2D + 3LDD_{ff}
    \end{equation}
    which evaluates to approximately $8.83 \times 10^8$ multiplications. To summarize, in a forward pass through the network, the client performs approx. \underline{$\textbf{12.66}$\%} of the entire operations.        
\end{itemize}

Once the forward pass is complete, the gradient backpropagation through Split-2 is handled by the SP, and through Split-1 by the GS. The client only participates in computing the loss and initiating the backward pass. Typically, the backward pass incurs $2–3\times$ the cost of the forward pass due to additional gradient calculations, activation derivatives, and layer-specific operations (e.g., dropout, batch normalization). Taking a conservative estimate of $2.5\times$ the forward pass cost, the client's overall contribution in one full training iteration (forward + backward) comes out to be approximately \underline{$\textbf{5}$\%} of the total operations. 

To summarize, In a single training round, the smart meter (client) is responsible for:
\begin{itemize}
    \item $\approx 4.55$\% of the total communication overhead,
    \item $\approx 5$\% of the total computational overhead.
\end{itemize} 

\subsection{Latency Overhead} \label{app:A3}
As discussed earlier, the client handles approximately $12.66$\% of the total operations during the forward pass. Similarly, during the backward pass, the GS processes an equivalent portion, with the remainder executed by the SP. Therefore, the end-to-end latency is predominantly influenced by the utility provider's infrastructure (GS and SP entities).

This latency can be effectively minimized by deploying high-performance hardware, particularly CUDA-enabled GPUs, which offer significant acceleration over even high-end CPUs for matrix-heavy operations typical in deep learning. As reported in Section IV-A of the manuscript, a single training epoch takes approximately $20$ minutes for both \textit{SplitGlobal} and \textit{SplitPersonal} frameworks, and $40$ minutes for the \textit{Central} model.

It's important to note that our implementation was conducted on a single machine, without deploying actual smart meters and separate GS-SP hardware. Thus, while we cannot provide a real-world latency measurement, the breakdown presented above offers a framework for estimating latency under practical deployment scenarios.

To illustrate this, consider the following assumptions:
\begin{itemize}
    \item The GS-SP setup performs multiplication operations $5×$ faster than the client device.
    \item The GS-SP together are responsible for $95$\% of the operations and can complete these in $10$ seconds.
\end{itemize}
Under these assumptions, the entire round of model update (if only performed by GS-SP) would take $10/0.95 \approx 10.53$ seconds. Consequently, the client's $5$\% share would take $0.053\times 5 \approx 2.64$ seconds.

Hence, the total latency per round would be approximately $12.64$ seconds, with the client-side accounting for around $21$\% of the total latency.

\subsection{Energy Overhead} \label{app:A4}
Building on the previous discussions, and assuming identical hardware architecture across the client, GS, and SP systems, the client would be responsible for only around $5$\% of the total energy consumption during a single model training round, while the remaining $95$\% would be incurred by the GS-SP infrastructure.

In our system model, the utility provider manages both the GS and SP components, allowing us to consider the architecture as being divided into two principal entities: the provider and the client. Within this setting, offloading just $5$\% of the total computation to the client achieves a significant accuracy gain. As shown in Table II of the revised manuscript, the SplitPersonal model achieves a mean prediction MSE of $0.251$, improving upon the central model’s MSE of $0.308$, which translates to an $18.51$\% performance improvement.

If we temporarily disregard the additional communication and orchestration overhead introduced by the split learning framework and assume uniform energy usage per operation across devices, this results in an $18.51$\% gain in performance at effectively no extra energy cost to the client. While this is an idealized scenario, it highlights the efficiency potential of the proposed architecture, even when only a minimal portion of the workload is handled by the client device.
It is also worth noting that in practical deployments, client devices such as smart meters may be equipped with energy-efficient but lower-performance hardware. In such cases, even though the latency of the client-side computation may increase, the overall energy consumption per operation could be significantly lower compared to high-performance GS-SP infrastructure. This further supports the viability of our split learning approach, as the computationally light client-side load (only around $5$\%) could be executed with minimal energy footprint, without compromising model performance.

%================================================================================================
\begin{IEEEbiography}[{\includegraphics[width=1in,height=1.25in,clip,keepaspectratio]{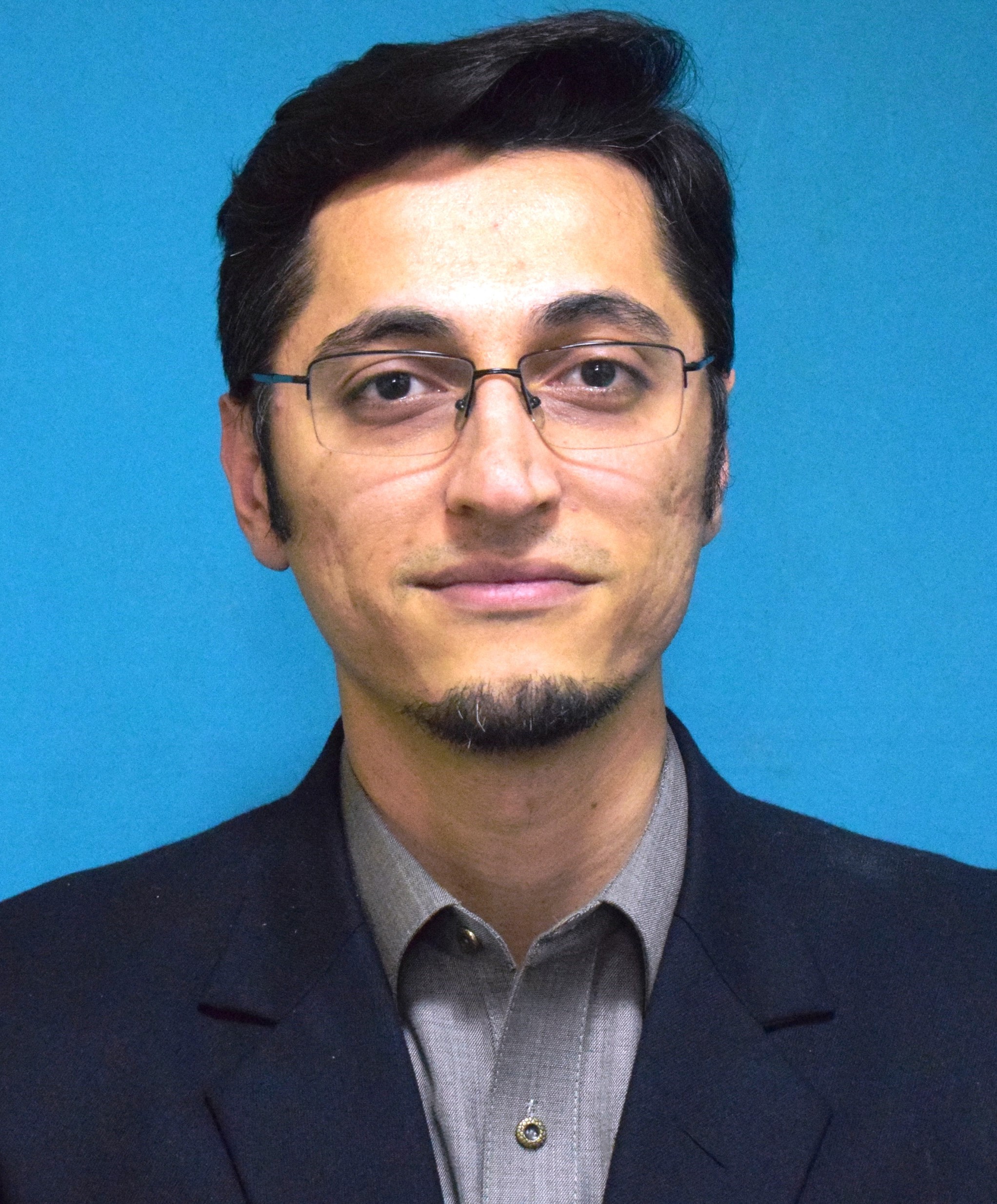}}]{Asif Iqbal} received the BS degree in Telecommunication Engineering from NUCES-FAST, Peshawar, Pakistan, MS degree in Wireless Communications from LTH, Lunds University, Sweden, and Ph.D in Electrical \& Electronics Engineering from The University of Melbourne, Melbourne, Australia in 2008, 2011, and 2019 respectively. 
He is currently working as a Research Fellow with the Department of Electrical \& Computer Engineering at the National University of Singapore, Singapore. Dr. Iqbal previously served on the faculty of NUCES-FAST, Peshawar, Pakistan as an Assistant Professor. His research interests include signal processing, deep learning, sparse signal representations, and privacy-preserving machine learning.
\end{IEEEbiography}
% %------------------------------
\vskip -2\baselineskip plus -1fil
 \begin{IEEEbiography}[{\includegraphics[width=1.12in,height=1.35in,clip,keepaspectratio]{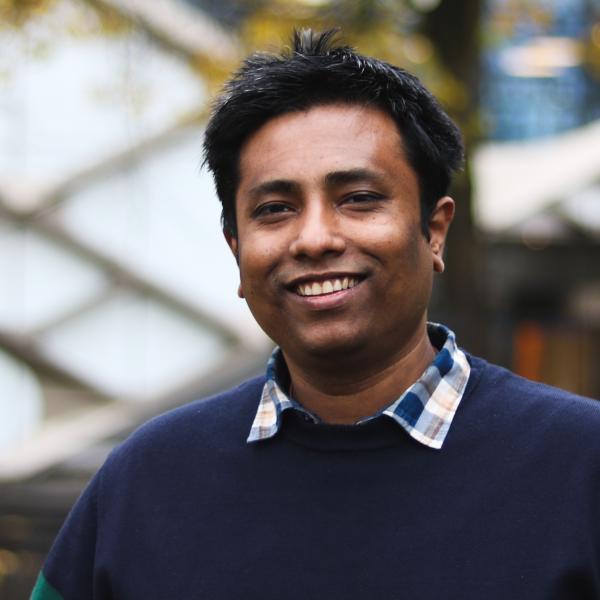}}]{Prosanta Gope} (Senior Member, IEEE) is currently working as an Associate Professor in the Department of Computer Science (Cyber Security) at the University of Sheffield, UK. Dr Gope served as a Research Fellow in the Department of Computer Science at the National University of Singapore (NUS). Primarily driven by tackling challenging real-world security problems, he has expertise in Lightweight Authentication, Authenticated Encryption, 5G and Next Generation Communication Security, Privacy-Preserving Machine Learning, Security in the Internet of Things, Smart-Grid Security,  PUF-based
security system
and IoT Hardware. He has authored more than 100 peer-reviewed articles in several reputable international journals and conferences and has four filed patents. Several of his papers have been published in high-impact journals (such as IEEE TIFS, IEEE TDSC, and IEEE/ACM TON), and prominent security conferences (such as IEEE S\&P,  ACM CCS, IEEE Computer Security Foundations Symposium (CSF), Privacy Enhancing Technologies Symposium (PETS), ESORICS, Euro S\&P, IEEE TrustCom, IEEE HoST, etc.) Dr Gope has been a TPC member and Co-Chair in several reputable international conferences, including PETS, ESORICS, IEEE TrustCom, IEEE GLOBECOM (Security Track), and ARES. He currently serves as an Associate Editor of the IEEE Transactions on Dependable and Secure Computing, IEEE Transactions on Information \& Forensics  Security, IEEE Transactions on Services Computing, IEEE Systems Journal, and the Journal of Information Security and Applications (Elsevier). His research has been funded by EPSRC, Innovate UK, and the Royal Society.
\end{IEEEbiography}
% %---------------------------------
\vskip -2\baselineskip plus -1fil
\begin{IEEEbiography}[{\includegraphics[width=1in,height=1.25in,clip,keepaspectratio]{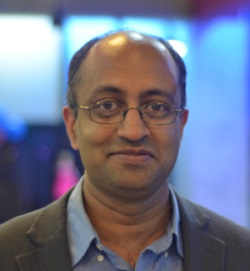}}]{Biplab Sikdar} \textnormal{(S’98-M’02-SM’09)} received the B.Tech. Degree in Electronics and Communication Engineering from North Eastern Hill University, Shillong, India, in 1996, and the M.Tech. Degree in electrical engineering from the Indian Institute of Technology Kanpur, Kanpur, India, in 1998, and the Ph.D. degree in electrical engineering from the Rensselaer Polytechnic Institute, Troy, NY, USA, in 2001. He was a faculty member at the Rensselaer Polytechnic Institute from 2001 to 2013 and was an assistant professor and Associate Professor. He is currently a Professor and Head of the Department of Electrical and Computer Engineering at the National University of Singapore. He also serves as the Director of the Cisco-NUS Corporate Research Laboratory. His current research interests include wireless networks and security for the Internet of Things and cyber-physical systems. He has served as an Associate Editor for the IEEE Transactions on Communications, IEEE Transactions on Mobile Computing, IEEE Internet of Things Journal and IEEE Open Journal of Vehicular Technology.
 \end{IEEEbiography}
%================================================================================================
% \newpage 
\end{document}